\RequirePackage{fix-cm}
\documentclass[twocolumn]{svjour3}           
\smartqed  
\usepackage{graphicx}
\usepackage{wrapfig}
\usepackage{multirow} 
\usepackage[utf8x]{inputenc}
\usepackage{pgfplots}
\usepackage{pgfplotstable}
\pgfplotsset{compat=newest}
\usepackage{tikz}
\usetikzlibrary{backgrounds}
\usepackage{amsmath}
\usepackage{amssymb}
\usepackage{color}
\usepackage{xcolor}
\usepackage{booktabs,siunitx} 
\usepackage{makecell}
\usepackage{enumitem}
\usepackage[caption=false]{subfig}
\usepackage[noadjust]{cite}
\usepackage{array,multirow,graphicx}
\usepackage[colorlinks=true,allcolors=green]{hyperref}
\usepackage{rotating}

\newcommand{\ie}{\mbox{\emph{i.e.\ }}}
\newcommand{\eg}{\mbox{\emph{e.g.\ }}}
\newcommand{\etal}{\mbox{\emph{et al.}}}

\DeclareMathOperator*{\Card}{card}
\DeclareMathOperator*{\Median}{median}

\newcommand{\best}[1]{\textbf{#1}}

\newcommand{\fnn}[1]{$\scriptstyle^{#1}$}

\definecolor{dcn}                 {RGB}{227,111, 38}
\definecolor{dcn-ft-0005}         {RGB}{128, 64,128}
\definecolor{dcn-ft-001}          {RGB}{ 62,135,207}
\definecolor{dcn-ft-002}          {RGB}{190,153,153}
\definecolor{dcn-ft-003}          {RGB}{107,142, 35}
\definecolor{dcn-ft-006}          {RGB}{200,  0,  0}

\definecolor{road}                {RGB}{128, 64,128}
\definecolor{sidewalk}            {RGB}{244, 35,232}
\definecolor{building}            {RGB}{ 70, 70, 70}
\definecolor{wall}                {RGB}{102,102,156}
\definecolor{fence}               {RGB}{190,153,153}
\definecolor{pole}                {RGB}{153,153,153}
\definecolor{traffic light}       {RGB}{250,170, 30}
\definecolor{traffic sign}        {RGB}{220,220,  0}
\definecolor{vegetation}          {RGB}{107,142, 35}
\definecolor{terrain}             {RGB}{152,251,152}
\definecolor{sky}                 {RGB}{ 70,130,180}
\definecolor{person}              {RGB}{220, 20, 60}
\definecolor{rider}               {RGB}{255,  0,  0}
\definecolor{car}                 {RGB}{  0,  0,142}
\definecolor{truck}               {RGB}{  0,  0, 70}
\definecolor{bus}                 {RGB}{  0, 60,100}
\definecolor{train}               {RGB}{  0, 80,100}
\definecolor{motorcycle}          {RGB}{  0,  0,230}
\definecolor{bicycle}             {RGB}{119, 11, 32}
\definecolor{void}                {RGB}{  0,  0,  0}

\definecolor{fine}{RGB}{62,135,207}
\definecolor{coarse}{RGB}{220,20,60}
\begin{document}

\title{Semantic Foggy Scene Understanding with Synthetic Data
}

\author{Christos Sakaridis \and Dengxin Dai \and Luc Van Gool 
}


\institute{C. Sakaridis \and D. Dai \and L. Van Gool \at
              ETH Z\"urich, Zurich, Switzerland
           \and
           L. Van Gool \at
              KU Leuven, Leuven, Belgium
}

\date{Received: date / Accepted: date}

\maketitle

\begin{abstract}
\sloppy{This work addresses the problem of semantic foggy scene understanding (SFSU).
Although extensive research has been performed on image dehazing and on semantic scene understanding with clear-weather images, little attention has been paid to SFSU. 
Due to the difficulty of collecting and annotating foggy images, we choose to generate synthetic fog on real images that depict clear-weather outdoor scenes, and then leverage these partially synthetic data for SFSU by employing state-of-the-art convolutional neural networks (CNN). In particular, a complete pipeline to add synthetic fog to real, clear-weather images using incomplete depth information is developed. We apply our fog synthesis on the Cityscapes dataset and generate \emph{Foggy Cityscapes} with 20550 images. SFSU is tackled in two ways: 1) with typical supervised learning, and 2) with a novel type of semi-supervised learning, which combines 1) with an unsupervised supervision transfer from clear-weather images to their synthetic foggy counterparts. In addition, we carefully study the usefulness of image dehazing for SFSU. For evaluation, we present \emph{Foggy Driving}, a dataset with 101 real-world images depicting foggy driving scenes, which come with ground truth annotations for semantic segmentation and object detection. 
Extensive experiments show that 1) supervised learning with our synthetic data significantly improves the performance of state-of-the-art CNN for SFSU on \emph{Foggy Driving}; 2) our semi-supervised learning strategy further improves performance; and 3) image dehazing marginally advances SFSU with our learning strategy.  The datasets, models and code are made publicly available.}
\end{abstract}

\section{Introduction}
\label{sec:intro}

\begin{figure*}
\centering
\includegraphics[width=0.9\textwidth]{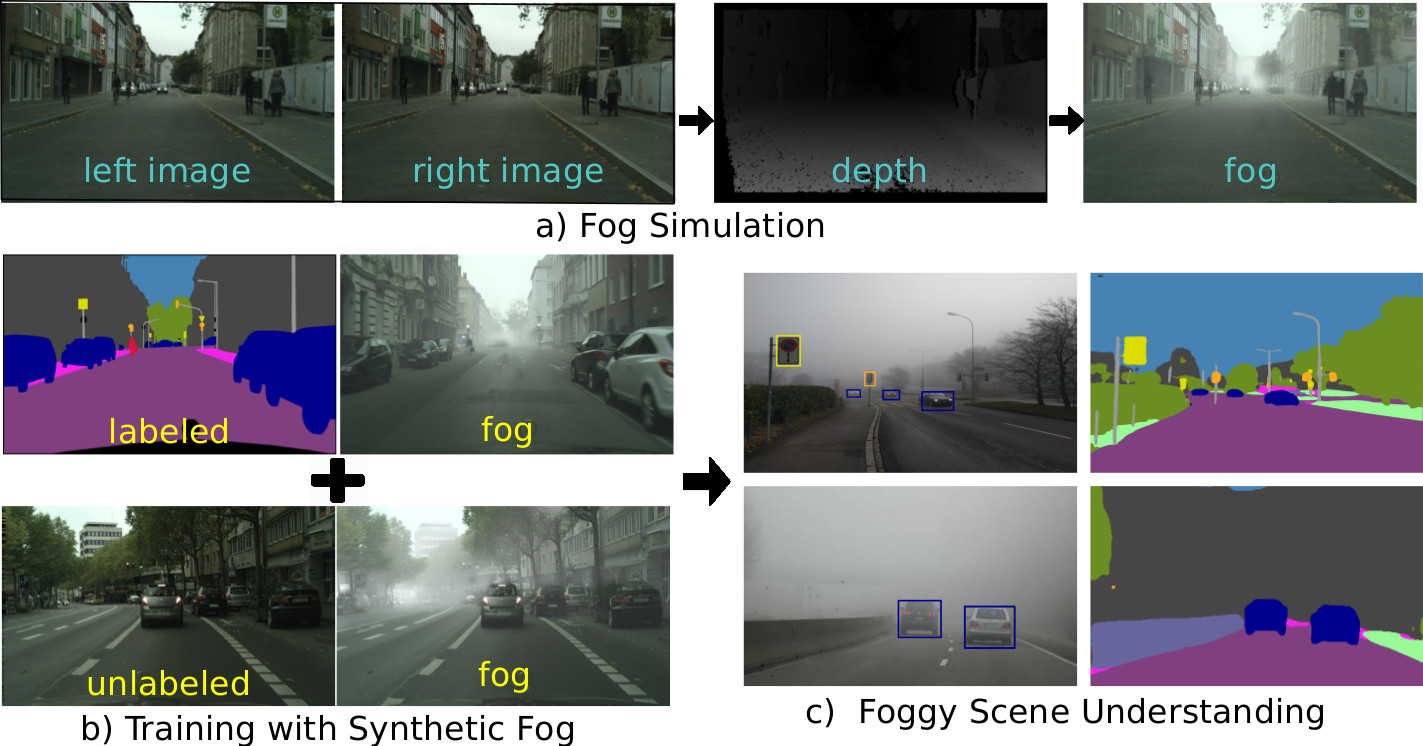}  
\caption{The pipeline of semantic foggy scene understanding with partially synthetic data: from a) fog simulation on real outdoor scenes, to b) training with pairs of such partially synthetic foggy images and semantic annotations as well as pairs of foggy images and clear-weather images, and c) scene understanding of real foggy scenes. This figure is seen better on a screen
} 
\label{fig:pipeline}
\end{figure*}

\sloppy{Cameras and the accompanying vision algorithms are widely used for applications such as surveillance~\cite{cv4traffic}, remote sensing~\cite{dai:satellite:image}, and automated cars~\cite{cv4av:17}, and their deployment keeps expanding. While these sensors and algorithms are constantly getting better, they are mainly designed to operate on clear-weather images and videos~\cite{vision:atmosphere}. Yet, outdoor applications can hardly escape from ``bad'' weather. Thus, such computer vision systems should also function under adverse weather conditions. Here we focus on the presence of fog.}

Fog degrades the visibility of a scene significantly~\cite{contrast:weather:degraded,tan2008visibility}. This causes problems not only to human observers, but also to computer vision algorithms. During the past years, a large body of research has been conducted on image defogging (dehazing) to increase scene visibility~\cite{bayesian:defogging,dark:channel,dehazing:mscale:depth}. Meanwhile, marked progress has been made in semantic scene understanding with clear-weather images and videos~\cite{faster:rcnn,Cityscapes,dilated:convolution}. In contrast, the semantic understanding of foggy scenes has received little attention, despite its importance in outdoor applications. For instance, an automated car still requires a robust detection of road lanes, traffic lights, and other traffic agents in the presence of fog. This work investigates semantic foggy scene understanding (SFSU).

High-level semantic scene understanding is usually tackled by learning from many annotations of real images~\cite{imagenet:2015,Cityscapes}. Yet, the difficulty of collecting and annotating images for unusual weather conditions such as fog renders this standard protocol problematic. To overcome this problem, we depart from this traditional paradigm and propose another route, also different from moving to fully synthetic scenes. Instead, we choose to generate synthetic fog into real images that contain clear-weather outdoor scenes, and then leverage these partially synthetic foggy images for SFSU.

Given the fact that large-scale annotated data are available for clear-weather images~\cite{pascal:2011,kitti,Cityscapes,imagenet:2015}, we present an automatic pipeline to add synthetic yet highly realistic fog to such datasets. Our fog simulation uses the standard optical model for daytime fog~\cite{Koschmieder:optical:model} (which has already been used extensively in image dehazing) to overlay existing clear-weather images with synthetic fog in a physically sound way, simulating the underlying mechanism of foggy image formation. We leverage our fog simulation pipeline to create our \emph{Foggy Cityscapes} dataset, by adding fog to urban scenes from the Cityscapes dataset~\cite{Cityscapes}. This has led to 550 carefully refined high-quality synthetic foggy images with fine semantic annotations inherited directly from Cityscapes, plus an additional 20000 synthetic foggy images without fine annotations. The resulting ``synthetic-fog'' images are used to adapt two semantic segmentation models~\cite{dilated:convolution,refinenet} and an object detector~\cite{fast:rcnn} to foggy scenes. The models are trained in two fashions: 1) by the typical supervised learning scheme, using the 550 high-quality annotated foggy images, and 2) by a novel semi-supervised learning approach, which augments the dataset that is used in 1) with the additional 20000 foggy images and draws the missing supervision for these images from the predictions of the source, clear-weather model on their clear-weather counterparts. For evaluation purposes, we collect and annotate a new dataset, \emph{Foggy Driving}, with 101 images of driving scenes in the presence of fog. See \autoref{fig:pipeline} for the whole pipeline of our work. In addition, this work studies the utility of three state-of-the-art image dehazing methods for SFSU as well as human understanding of foggy scenes.

The main contributions of the paper are: 1) an automatic and scalable pipeline to impose high-quality synthetic fog on real clear-weather images; 2) two new datasets, one synthetic and one real, to facilitate training and evaluation of models used in SFSU; 3) a new semi-supervised learning approach for SFSU; and 4) a detailed study of the benefit of image dehazing for SFSU and human perception of foggy scenes. 

\sloppy{The rest of the paper is organized as follows. Section~\ref{sec:related} presents the related work. Section~\ref{sec:fog:simulation} is devoted to our fog simulation pipeline, followed by Section~\ref{sec:datasets} that introduces our two foggy datasets. Section~\ref{sec:learning:sl} describes supervised learning with our synthetic foggy data and studies the usefulness of image dehazing for SFSU in this context. Finally, Section~\ref{sec:learning:ssl} extends the learning to a semi-supervised paradigm, where supervision is transferred from clear-weather images to their synthetic foggy counterparts, and Section~\ref{sec:conclusion} concludes the paper.}

\section{Related Work}
\label{sec:related} 

Our work is relevant to image defogging (dehazing), depth denoising and completion, foggy scene understanding, synthetic visual data, and transfer learning.

\subsection{Image Defogging/Dehazing} 
Fog fades the color of observed objects and reduces their contrast. 
Extensive research has been conducted on image defogging (dehazing) to increase the visibility of foggy scenes. This ill-posed problem has been tackled from different perspectives. For instance, in contrast enhancement~\cite{contrast:weather:degraded,tan2008visibility} the rationale is that clear-weather images have higher contrast than images degraded by fog. Depth and statistics of natural images are exploited as priors as well~\cite{bayesian:defogging,fattal2008single,nonlocal:image:dehazing,fattal2014dehazing}. Another line of work is based on the dark channel prior~\cite{dark:channel}, with the empirically validated assumption that pixels of clear-weather images are very likely to have low values in some of the three color channels. Certain works focus particularly on enhancing foggy road scenes~\cite{THC+12,exponential:contrast:restoration}. Methods have also been developed for nighttime~\cite{LTB15}, given its importance in outdoor applications. Fast dehazing approaches have been developed in~\cite{fast:visibility:restoration:09,fast:dehazing:linear:transform} towards real-time applications. Recent approaches also rely on trainable architectures~\cite{TYW14}, which have evolved to end-to-end models~\cite{RLZ+16,joint:transmission:estimation:dehazing,deep:transmission:network}. For a comprehensive overview of defogging/dehazing algorithms, we point the reader to~\cite{review:defogging:restoration,dehazing:survey:benchmarking}. All these approaches can greatly increase visibility. Our work is complementary and focuses on the semantic understanding of foggy scenes.

\subsection{Depth Denoising and Completion}
Synthesizing a foggy image from its real, clear counterpart generally requires an accurate depth map. In previous works, the colorization approach of~\cite{colorization} has been used to inpaint depth maps of the \emph{indoor} NYU Depth dataset~\cite{nyu:depth}. Such inpainted depth maps have been used in state-of-the-art dehazing approaches such as~\cite{RLZ+16} to generate training data in the form of synthetic indoor foggy images. In contrast, our work considers real \emph{outdoor} urban scenes from the Cityscapes dataset~\cite{Cityscapes}, which contains significantly more complex depth configurations than NYU Depth. Furthermore, the available depth information in Cityscapes is not provided by a depth sensor, but it is rather an estimate of the depth resulting from the application of a semiglobal matching stereo algorithm based on~\cite{sgm}. This depth estimate usually contains a large amount of severe artifacts and large holes (cf.\ \autoref{fig:pipeline}), which render it inappropriate for direct use in fog simulation. There are several recent approaches that handle highly noisy and incomplete depth maps, including stereoscopic inpainting~\cite{stereoscopic:inpainting}, spatio-temporal hole filling~\cite{spatiotemporal:hole:filling} and layer depth denoising and completion~\cite{layer:depth:denoising}. Our method builds on the framework of stereoscopic inpainting~\cite{stereoscopic:inpainting} which performs depth completion at the level of superpixels, and introduces a novel, theoretically grounded objective for the superpixel-matching optimization that is involved.

\subsection{Foggy Scene Understanding}
Semantic understanding of outdoor scenes is a crucial enabler for applications such as assisted or autonomous driving. Typical examples include road and lane detection~\cite{recent:progress:lane}, traffic light detection~\cite{traffic:light:survey:16}, car and pedestrian detection~\cite{kitti}, and a dense, pixel-level segmentation of road scenes into most of the relevant semantic classes~\cite{recognition:sfm:eccv08,Cityscapes}. While deep recognition networks have been developed~\cite{dilated:convolution,refinenet,pspnet,fast:rcnn,faster:rcnn} and large-scale datasets have been presented~\cite{kitti,Cityscapes}, that research mainly focused on clear weather. There is also a large body of work on fog detection~\cite{fog:detection:cv:09,fog:detection:vehicles:12,night:fog:detection,fast:fog:detection}. Classification of scenes into foggy and fog-free has been tackled as well~\cite{fog:nonfog:classification:13}. In addition, visibility estimation has been extensively studied for both daytime~\cite{visibility:road:fog:10,visibility:detection:fog:15,fog:detection:visibility:distance} and nighttime~\cite{night:visibility:analysis:15}, in the context of assisted and autonomous driving. The closest of these works to ours is~\cite{visibility:road:fog:10}, in which synthetic fog is generated and foggy images are segmented to \emph{free-space area} and \emph{vertical objects}. Our work differs in that: 1) our semantic understanding task is more complex, with $19$ semantic classes that are commonly involved in driving scenarios, 8 of which occur as distinct objects; 2) we tackle the problem with modern deep CNN for semantic segmentation~\cite{dilated:convolution,refinenet} and object detection~\cite{fast:rcnn}, taking full advantage of the most recent advances in this field; and 3) we compile and release a large-scale dataset of synthetic foggy images based on real scenes plus a dataset of real-world foggy scenes, featuring both dense pixel-level semantic annotations and annotations for object detection.

\subsection{Synthetic Visual Data}
\sloppy{The leap of computer vision in recent years can to an important extent be attributed to the availability of large, labeled datasets~\cite{pascal:2011,imagenet:2015,Cityscapes}. However, acquiring and annotating such a dataset for each new problem is not (yet) doable. Thus, learning with synthetic data is gaining attention. We give some notable examples. Dosovitskiy \etal~\cite{flow:chair} use the renderings of a floating chair to train dense optical flow regression networks. Gupta \etal~\cite{synthetic:text} impose text onto natural images to learn an end-to-end text detection system. V{\'a}zquez \etal~\cite{virtual:pedestrian} train pedestrian detectors with virtual data. In \cite{Synthia:dataset,playing:data} the authors leverage video game engines to render images along with dense semantic annotations that are subsequently used in combination with real data to improve the semantic segmentation performance of modern CNN architectures on real scenes. Going one step further, \cite{driving:in:the:matrix} shows that for the task of vehicle detection, training a CNN model \emph{only} on massive amounts of synthetic images can outperform the same model trained on large-scale real datasets like Cityscapes. By contrast, our work tackles semantic segmentation and object detection for real \emph{foggy} urban scenes, by adding synthetic fog to \emph{real} images taken under clear weather. Hence, our approach is based on only partially synthetic data. In the same vein, \cite{augmented:reality:car:segmentation} is based on real urban scenes, augmented with virtual cars. A very interesting project is ``FOG''~\cite{artificial:fog:production:08}. Its team developed a prototype of a small-scale fog chamber, able to produce stable visibility levels and homogeneous fog to test the reaction of drivers.}

\subsection{Transfer Learning}
\sloppy{Our work bears resemblance to works from the broad field of transfer learning. Model adaptation across weather conditions to semantically segment simple road scenes is studied in~\cite{road:scene:2013}. More recently, a domain adversarial based approach was proposed to adapt semantic segmentation models both at pixel level and feature level from simulated to real environments~\cite{cyCADA}. Our work generates synthetic fog from clear-weather data to close the domain gap. Combining our method and the aforementioned transfer learning methods is a promising direction for future work. The supervision transfer from clear weather to foggy weather in this paper is inspired by the stream of work on model distillation/imitation~\cite{hinton2015distilling,supervision:transfer,dai:metric:imitation}. Our approach is similar in that knowledge is transferred from one domain (model) to another by using paired data samples as a bridge.}

\section{Fog Simulation on Real Outdoor Scenes}
\label{sec:fog:simulation}

To simulate fog on input images that depict real scenes with clear weather, the standard approach is to model the effect of fog as a function that maps the radiance of the clear scene to the radiance observed at the camera sensor. Critically, this space-variant function is usually parameterized by the distance $\ell$ of the scene from the camera, which equals the length of the path along which light has traveled and is closely related to scene depth. As a result, the pair of the clear image and its depth map forms the basis of our foggy image synthesis. In this section, we first detail the optical model which we use for fog and then present our complete pipeline for fog simulation, with emphasis on our denoising and completion of the input depth. Finally, we present some criteria for selecting suitable images to generate high-quality synthetic fog.

\subsection{Optical Model of Choice for Fog}
\label{sec:fog:simulation:model}

\sloppy{In the image dehazing literature, various optical models have been used to model the effect of haze on the appearance of a scene. For instance, optical models tailored for nighttime haze removal have been proposed in~\cite{nighttime:dehazing:imaging:model,LTB15}, taking into account the space-variant lighting that characterizes most nighttime scenes. This variety of models is directly applicable to the case of fog as well, since the physical process for image formation in the presence of either haze or fog is essentially similar. For our synthesis of foggy images, we consider the standard optical model of~\cite{Koschmieder:optical:model}, which is used extensively in the literature \cite{dark:channel,fattal2008single,TYW14,fast:visibility:restoration:09,RLZ+16} and is formulated as}
\begin{equation} \label{eq:haze:model}
\vec{I}(\vec{x}) = \vec{R}(\vec{x})t(\vec{x}) + \vec{L}(1 - t(\vec{x})),
\end{equation}
where $\vec{I}(\vec{x})$ is the observed foggy image at pixel $\vec{x}$, $\vec{R}(\vec{x})$ is the clear scene radiance and $\vec{L}$ is the atmospheric light. This model assumes the atmospheric light to be globally constant, which is generally valid only for \textit{daytime} images. The transmission $t(\vec{x})$ determines the amount of scene radiance that reaches the camera. In case of a \textit{homogeneous} medium, transmission depends on the distance $\ell(\vec{x})$ of the scene from the camera through
\begin{equation} \label{eq:transmission}
t(\vec{x}) = \exp\left(-\beta\ell(\vec{x})\right).
\end{equation}
The parameter $\beta$ is named attenuation coefficient and it effectively controls the thickness of the fog: larger values of $\beta$ mean thicker fog. The meteorological optical range (MOR), also known as visibility, is defined as the maximum distance from the camera for which $t(\vec{x}) \geq 0.05$, which implies that if \eqref{eq:transmission} is valid, then $\text{MOR}=2.996/\beta$. Fog decreases the MOR to less than 1 km by definition~\cite{Federal:meteorological:handbook}. Therefore, the attenuation coefficient in homogeneous fog is by definition
\begin{equation} \label{eq:beta:bound:fog}
\beta \geq 2.996\times{}10^{-3}{\text{ m}}^{-1},
\end{equation}
where the lower bound corresponds to the lightest fog configuration. In our fog simulation, the value that is used for $\beta$ always obeys \eqref{eq:beta:bound:fog}.

\begin{figure*}[tb]
    \centering
    \subfloat{\includegraphics[width=0.325\textwidth]{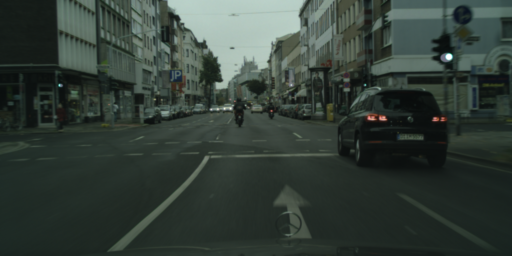}}
    \hfil
    \subfloat{\includegraphics[width=0.325\textwidth]{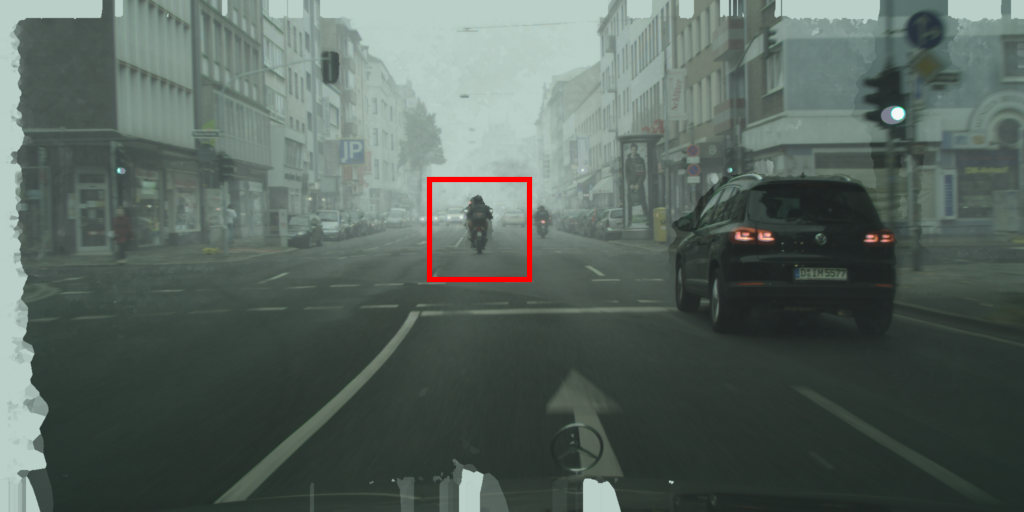}}
    \hfil
    \subfloat{\includegraphics[width=0.325\textwidth]{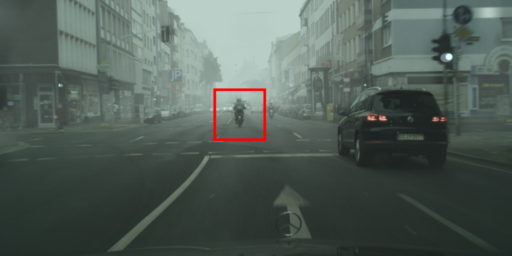}}
    \\
    \addtocounter{subfigure}{-3}
    \subfloat[Input from Cityscapes]{\includegraphics[width=0.325\textwidth]{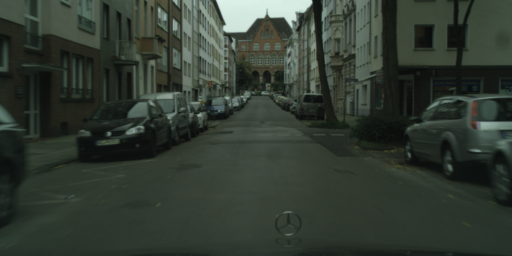}
    \label{fig:fog:simulation:input}}
    \hfil
    \subfloat[Nearest-neighbor depth completion]{\includegraphics[width=0.325\textwidth]{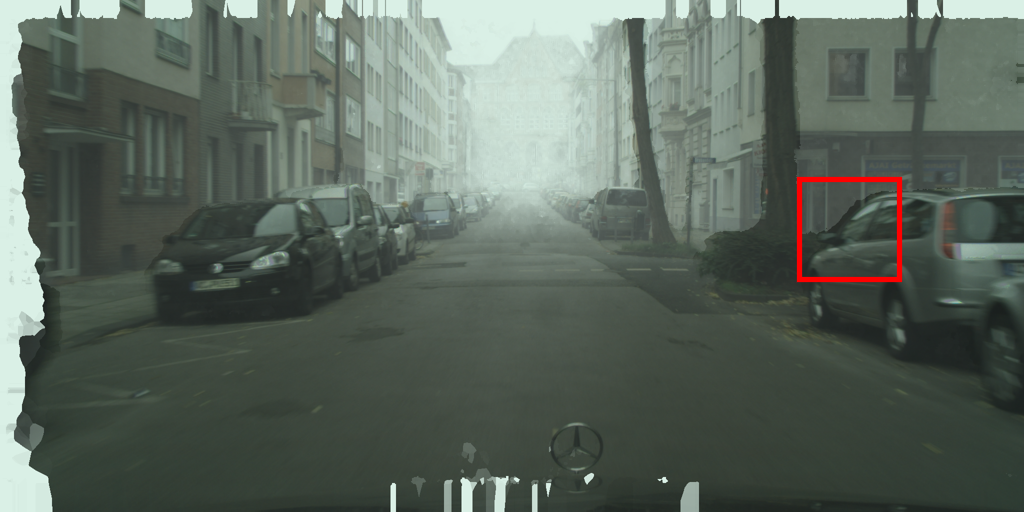}
    \label{fig:fog:simulation:nearest}}
    \hfil
    \subfloat[Our fog simulation]{\includegraphics[width=0.325\textwidth]{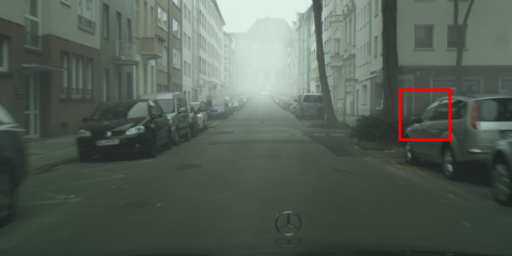}
    \label{fig:fog:simulation:ours}}
    \caption{Comparison of our fog simulation to nearest-neighbor interpolation for depth completion on images from Cityscapes. This figure is better seen on a screen and zoomed in}
    \label{fig:fog:simulation}
\end{figure*}

Model \eqref{eq:haze:model} provides a powerful basis for simulating fog on outdoor scenes with clear weather. Even though its assumption of homogeneous atmosphere is strong, it generates synthetic foggy images that can act as good proxies for real world foggy images where this assumption might not hold exactly, as long as it is provided with an \emph{accurate} transmission map $t$. Straightforward extensions of \eqref{eq:haze:model} are used in~\cite{THC+12} to simulate heterogeneous fog on synthetic scenes.

To sum up, the necessary inputs for fog simulation using \eqref{eq:haze:model} are a color image $\vec{R}$ of the original clear scene, atmospheric light $\vec{L}$ and a dense transmission map $t$ defined at each pixel of $\vec{R}$. Our task is thus twofold:
\begin{enumerate}
\item estimation of $t$, and \label{step:transmission}
\item estimation of $\vec{L}$ from $\vec{R}$. \label{step:light}
\end{enumerate}
Step~\ref{step:light} is simple: we use the method proposed in~\cite{dark:channel} with the improvement of~\cite{TYW14}. In the following, we focus on step~\ref{step:transmission} for the case of outdoor scenes with a noisy, incomplete estimate of depth serving as input.

\subsection{Depth Denoising and Completion for Outdoor Scenes}
\label{sec:fog:simulation:depth}

The inputs that our method requires for generating an accurate transmission map \(t\) are:
\begin{itemize}[label=\textbullet]
\item the original, clear-weather color image $\vec{R}$ to add synthetic fog on, which constitutes the \emph{left} image of a stereo pair,
\item the \emph{right} image $\vec{Q}$ of the stereo pair,
\item the intrinsic calibration parameters of the two cameras of the stereo pair as well as the length of the baseline,
\item a dense, raw disparity estimate $D$ for $\vec{R}$ of the same resolution as $\vec{R}$, and
\item a set $M$ comprising the pixels where the value of $D$ is missing.
\end{itemize}
These requirements can be easily fulfilled with a stereo camera and a standard stereo matching algorithm~\cite{sgm}. 

The main steps of our pipeline are the following:
\begin{enumerate}
\item calculation of a raw depth map $d$ in meters, \label{step:depth:raw}
\item \emph{denoising and completion} of $d$ to produce a refined depth map $d^{\prime}$ in meters, \label{step:depth:denoising}
\item calculation of a scene distance map $\ell$ in meters from $d^{\prime}$, \label{step:distance}
\item application of \eqref{eq:transmission} to obtain an initial transmission map $\hat{t}$, and \label{step:transmission:init}
\item guided filtering~\cite{guided:filtering} of $\hat{t}$ using $\vec{R}$ as guidance to compute the final transmission map $t$. \label{step:guided}
\end{enumerate}

The central idea is to leverage the accurate structure that is present in the color images of the stereo pair in order to improve the quality of depth, before using the latter as input for computing transmission. We now proceed in explaining each step in detail, except step~\ref{step:transmission:init} which is straightforward. In step~\ref{step:depth:raw}, we use the input disparity $D$ in combination with the values of the focal length and the baseline to obtain $d$. The missing values for $D$, indicated by $M$, are also missing in $d$.

Step~\ref{step:depth:denoising} follows a segmentation-based depth filling approach, which builds on the stereoscopic inpainting method presented in~\cite{stereoscopic:inpainting}. More specifically, we use a superpixel segmentation of the clear image $\vec{R}$ to guide depth denoising and completion at the level of superpixels, making the assumption that each individual superpixel corresponds roughly to a plane in the 3D scene.

First, we apply a photo-consistency check between $\vec{R}$ and $\vec{Q}$, using the input disparity $D$ to establish pixel correspondences between the two images of the stereo pair, similar to Equation~(12) in~\cite{stereoscopic:inpainting}. All pixels in $\vec{R}$ for which the color deviation (measured as difference in the RGB color space) from the corresponding pixel in $\vec{Q}$ has greater magnitude than $\epsilon = 12/255$ are deemed invalid regarding depth and hence are added to $M$.

We then segment $\vec{R}$ into superpixels with SLIC~\cite{slic:superpixels}, denoting the target number of superpixels as $\hat{K}$ and the relevant range domain scale parameter as $m = 10$. For depth denoising and completion on Cityscapes, we use $\hat{K} = 2048$. The final number of superpixels that are output by SLIC is denoted by $K$. These superpixels are classified into reliable and unreliable ones with respect to depth information, based on the number of pixels with missing or invalid depth that they contain. More formally, we use the criterion of Equation~(2) in~\cite{stereoscopic:inpainting}, which states that a superpixel $T$ is reliable if and only if
\begin{equation} \label{eq:reliable:criterion}
\Card(T\setminus{}M) \geq \max\{P,\,\lambda\Card(T)\},
\end{equation}
setting $P = 20$ and $\lambda = 0.6$.

For each superpixel that fulfills \eqref{eq:reliable:criterion}, we fit a depth plane by running RANSAC on its pixels that have a valid value for depth. We use an adaptive inlier threshold to account for differences in the range of depth values between distinct superpixels. For a superpixel $T$, the inlier threshold is set as
\begin{equation} \label{eq:ransac:inlier:thresh}
\theta = 0.01\Median_{\vec{x} \in T\setminus{}M}\{d(\vec{x})\}.
\end{equation}
We use adaptive RANSAC and set the maximum number of iterations to $2000$ and the bound on the probability of having obtained a pure inlier sample to $p = 0.99$.

The greedy approach of~\cite{stereoscopic:inpainting} is used subsequently to match unreliable superpixels to reliable ones pairwise and assign the fitted depth planes of the latter to the former. Different than~\cite{stereoscopic:inpainting}, we propose a novel objective function for matching pairs of superpixels. For a superpixel pair $(s,t)$, our proposed objective is formulated as
\begin{equation} \label{eq:superpixel:matching:objective}
E(s,t) = {\left\|\vec{C}_s - \vec{C}_t\right\|}^2 + \alpha {\left\|\vec{x}_s - \vec{x}_t\right\|}^2.
\end{equation}

The first term measures the proximity of the two superpixels in the range domain, where we denote the average CIELAB color of superpixel $s$ with $\vec{C}_s$. In other words, we penalize the squared Euclidean distance between the average colors of the superpixels in the CIELAB color space, which has been designed to increase perceptual uniformity~\cite{mean:shift}. On the contrary, the objective of~\cite{stereoscopic:inpainting} uses the cosine similarity of average superpixel colors to form the range domain cost:
\begin{equation} \label{eq:stereoscopic:inpainting:color:cost}
1-\frac{\vec{C}_s}{\|\vec{C}_s\|}\cdot\frac{\vec{C}_t}{\|\vec{C}_t\|}.
\end{equation}
The disadvantage of~\eqref{eq:stereoscopic:inpainting:color:cost} is that it assigns zero matching cost to dissimilar colors in certain cases. For instance, in the RGB color space, the pair of colors $(\delta,\delta,\delta)$ and $(1-\delta,1-\delta,1-\delta)$, where $\delta$ is a small positive constant, is assigned zero penalty, even though the former color is very dark gray and the latter is very light gray.

The second term on the right-hand side of~\eqref{eq:superpixel:matching:objective} measures the proximity of the two superpixels in the spatial domain as the squared Euclidean distance between their centroids $\vec{x}_s$ and $\vec{x}_t$. By contrast, the spatial proximity term of~\cite{stereoscopic:inpainting} assigns zero cost to pairs of adjacent superpixels and unit cost to non-adjacent pairs. This implies that close yet non-adjacent superpixels are penalized equally to very distant superpixels by~\cite{stereoscopic:inpainting}. As a result, a certain superpixel $s$ can be erroneously matched to a very distant superpixel $t$ which is highly unlikely to share the same depth plane as $s$, as long as the range domain term for this pair is minimal and all superpixels adjacent to $s$ are dissimilar to it with respect to appearance. Our proposed spatial cost handles these cases successfully: $t$ is assigned a very large spatial cost for being matched to $s$, and other superpixels that have less similar appearance yet smaller distance to $s$ are preferred.

In \eqref{eq:superpixel:matching:objective}, $\alpha > 0$ is a parameter that weights the relative importance of the spatial domain term versus the range domain term. Similarly to~\cite{slic:superpixels}, we set $\alpha = m^2/S^2$, where $S = \sqrt{N/K}$, $N$ denotes the total number of pixels in the image, and $m=10$ and $K$ are the same as for SLIC. Our matching objective~\eqref{eq:superpixel:matching:objective} is similar to the distance that is defined in SLIC~\cite{slic:superpixels} and other superpixel segmentation methods for assigning \emph{an individual pixel to a superpixel}. In our case though, this distance is rather used to measure similarity between \emph{pairs of superpixels}.

After all superpixels have been assigned a depth plane, we use these planes to complete the missing depth values for pixels belonging to $M$. In addition, we replace the depth values of pixels which do not belong to $M$ but constitute large-margin outliers with respect to their corresponding plane (deviation larger than $\hat{\theta} = 50\text{m}$) with the values imputed by the plane. This results in a complete, denoised depth map $d^{\prime}$, and concludes step~\ref{step:depth:denoising}.

In step~\ref{step:distance}, we compute the distance $\ell(\vec{x})$ of the scene from the camera at each pixel $\vec{x}$ based on $d^{\prime}(\vec{x})$, using the coordinates of the principal point plus the focal length of the camera.

Finally, in step~\ref{step:guided} we post-process the initial transmission map $\hat{t}$ with guided filtering~\cite{guided:filtering}, in order to smooth transmission while respecting the boundaries of the clear image $\vec{R}$. We fix the radius of the guided filter window to $r = 20$ and the regularization parameter to $\mu = 10^{-3}$, \ie{}we use the same values as in the haze removal experiments of~\cite{guided:filtering}.

Results of the presented pipeline for fog simulation on example images from Cityscapes are provided in \autoref{fig:fog:simulation} for $\beta = 0.01$, which corresponds to visibility of ca.\ $300\text{m}$. We compare our fog simulation to an alternative implementation, which employs nearest-neighbor interpolation to complete the missing values of the depth map before computing the transmission and does not involve guided filtering as a postprocessing step.

\subsection{Input Selection for High-Quality Fog Simulation}
\label{sec:fog:simulation:cityscapes}

Applying the presented pipeline to simulate fog on large datasets with real outdoor scenes such as Cityscapes with the aim of producing synthetic foggy images of high quality calls for careful refinement of the input.

\begin{figure}[tb]
    \centering
    \subfloat[Input image from Cityscapes]{\includegraphics[width=0.8\linewidth]{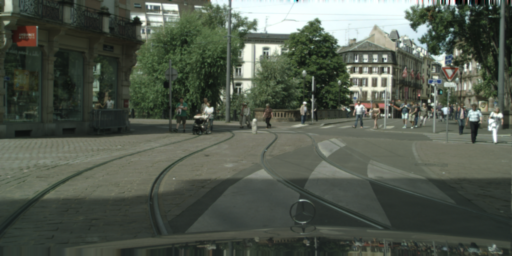}
    \label{fig:sunny:input}}\\
    \subfloat[Output of our fog simulation]{\includegraphics[width=0.8\linewidth]{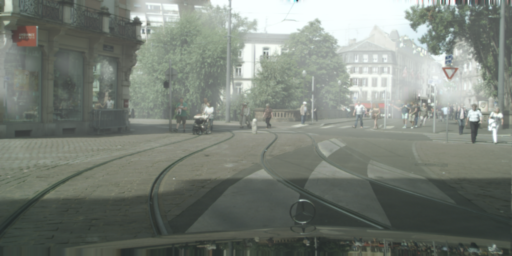}
    \label{fig:sunny:output}}
    \caption{Sunny scene from Cityscapes and the result of our fog simulation}
    \label{fig:sunny}
\end{figure}

To be more precise, the sky is clear in the majority of scenes in Cityscapes, with intense direct or indirect sunlight, as shown in \autoref{fig:sunny}\subref{fig:sunny:input}. These images usually contain sharp shadows and have high contrast compared to images that depict foggy scenes. This causes our fog simulation to generate synthetic images which do not resemble real fog very well, \eg{}\autoref{fig:sunny}\subref{fig:sunny:output}. Therefore, our first refinement criterion is whether the sky is overcast, ensuring that the light in the input real scene is not strongly directional.

Secondly, we observe that atmospheric light estimation in step~\ref{step:light} of our fog simulation sometimes fails to select a pixel with ground truth semantic label \emph{sky} as the representative of the value of atmospheric light. In rare cases, it even happens that the sky is not visible at all in an image. This results in an erroneous, physically invalid value of atmospheric light being used in \eqref{eq:haze:model} to synthesize the foggy image. Consequently, our second refinement criterion is whether the pixel that is selected as atmospheric light is labeled as \emph{sky}, and affords an automatic implementation.

\section{Foggy Datasets}
\label{sec:datasets}

We present two distinct datasets for semantic understanding of foggy scenes: \emph{Foggy Cityscapes} and \emph{Foggy Driving}. The former derives from the Cityscapes dataset~\cite{Cityscapes} and constitutes a collection of synthetic foggy images generated with our proposed fog simulation that automatically inherit the semantic annotations of their real, clear counterparts. On the other hand, \emph{Foggy Driving} is a collection of 101 real-world foggy road scenes with annotations for semantic segmentation and object detection, used as a benchmark for the domain of foggy weather.

\begin{figure*}[!tb]
\centering
$\begin{tabular}{cccc}
  \hspace{-2mm}
  \includegraphics[width=0.24\textwidth]{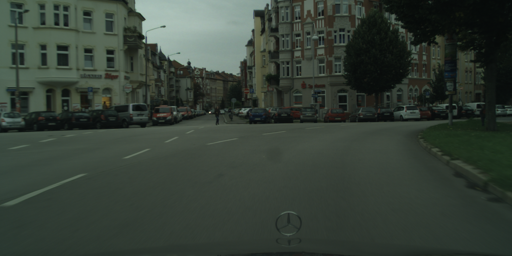} 
  & \hspace{-5mm}
  \includegraphics[width=0.24\textwidth]{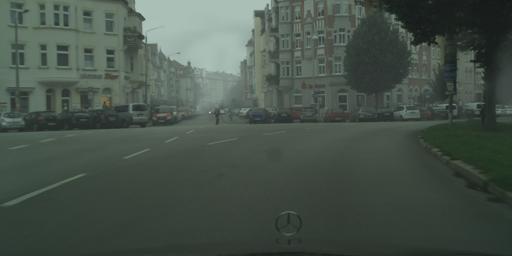}  
  & \hspace{-5mm}
  \includegraphics[width=0.24\textwidth]{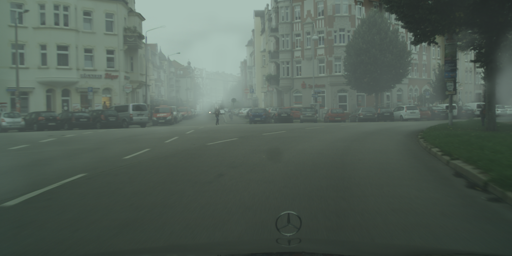}
  & \hspace{-5mm}
  \includegraphics[width=0.24\textwidth]{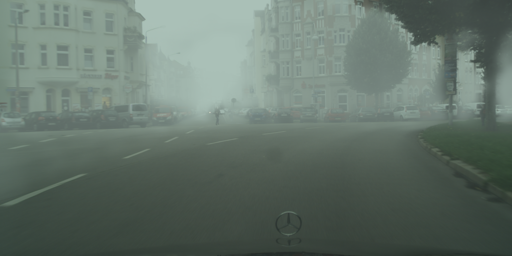}
 \\
 \hspace{-2mm}
\text{(a) clear-weather} & \hspace{-5mm}  \text{(b) $\beta=0.005$} & \hspace{-5mm} \text{(c) $\beta=0.01$} & \hspace{-5mm} \text{(d) $\beta=0.02$}\\
\end{tabular}$
\caption{Different versions of an exemplar scene from \emph{Foggy Cityscapes} for varying visibility}  
\label{fig:examples:pics} 
\end{figure*}

\subsection{Foggy Cityscapes}
\label{sec:datasets:foggy:cityscapes}

We apply the fog simulation pipeline that is presented in Section~\ref{sec:fog:simulation} to the complete set of images provided in the Cityscapes dataset. More specifically, we first obtain 20000 synthetic foggy images from the larger, coarsely annotated part of the dataset, and keep all of them, without applying the refinement criteria of Section~\ref{sec:fog:simulation:cityscapes}. In this way, we trade the high visual quality of the synthetic images for a very large scale and variability of the synthetic dataset. We do not make use of the original coarse annotations of these images for semantic segmentation; rather, we produce labellings with state-of-the-art semantic segmentation models on the original, clear images and use them to transfer knowledge from clear weather to foggy weather, as will be discussed in Section~\ref{sec:learning:ssl}. We name this set \emph{Foggy Cityscapes-coarse}.

In addition, we use the two criteria of Section~\ref{sec:fog:simulation:cityscapes} in conjunction to filter the finely annotated part of Cityscapes that originally comprises 2975 training and 500 validation images, and obtain a refined set of 550 images, 498 from the training set and 52 from the validation set, which fulfill both criteria. Running our fog simulation on this refined set provides us with a moderate-scale collection of high-quality synthetic foggy images. This collection automatically inherits the original fine annotations for \emph{semantic segmentation}, as well as bounding box annotations for \emph{object detection} which we generate by leveraging the instance-level semantic annotations that are provided in Cityscapes for the 8 classes \emph{person}, \emph{rider}, \emph{car}, \emph{truck}, \emph{bus}, \emph{train}, \emph{motorcycle} and \emph{bicycle}. We term this collection \emph{Foggy Cityscapes-refined}.

Since MOR can vary significantly in reality for different instances of fog, we generate five distinct versions of \emph{Foggy Cityscapes}, each of which is characterized by a constant simulated attenuation coefficient $\beta$ in \eqref{eq:transmission}, hence a constant MOR. In particular, we use $\beta \in \{0.005,\,0.01,\,0.02,\,0.03,\,0.06\}$, which correspond approximately to MOR of 600m, 300m, 150m, 100m and 50m respectively. \autoref{fig:examples:pics} shows three of the five synthesized foggy versions of a clear scene in \emph{Foggy Cityscapes}.

\subsection{Foggy Driving}
\label{sec:datasets:foggy:driving}

\emph{Foggy Driving} consists of 101 color images depicting real-world foggy driving scenes. We captured 51 of these images with a cell phone camera in foggy conditions at various areas of Zurich, and the rest 50 images were carefully collected from the web. We note that all images have been preprocessed so that they have a maximum resolution of $960\times{}1280$ pixels.

We provide dense, pixel-level semantic annotations for all images of \emph{Foggy Driving}. In particular, we use the 19 evaluation classes of Cityscapes: \emph{road}, \emph{sidewalk}, \emph{building}, \emph{wall}, \emph{fence}, \emph{pole}, \emph{traffic light}, \emph{traffic sign}, \emph{vegetation}, \emph{terrain}, \emph{sky}, \emph{person}, \emph{rider}, \emph{car}, \emph{truck}, \emph{bus}, \emph{train}, \emph{motorcycle} and \emph{bicycle}. Pixels that do not belong to any of the above classes or are not labeled are assigned the \emph{void} label, and they are ignored for semantic segmentation evaluation. At annotation time, we label individual instances of \emph{person}, \emph{rider}, \emph{car}, \emph{truck}, \emph{bus}, \emph{train}, \emph{motorcycle} and \emph{bicycle} separately following the Cityscapes annotation protocol, which directly affords bounding box annotations for these 8 classes.

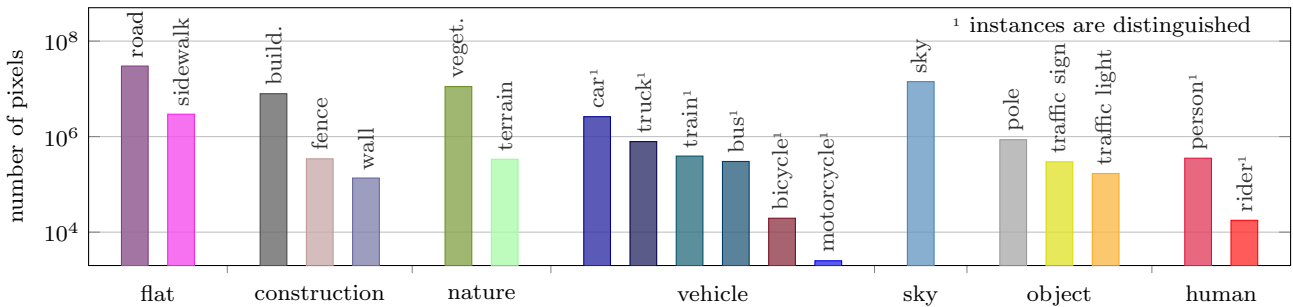
\begin{figure*}[tb]
    \centering
    \begin{tikzpicture}
    \tikzstyle{every node}=[font=\small]
    \begin{axis}[
            ybar,
            ymode=log,
            width=\textwidth,
            height=5cm,
            xmin=0,
            xmax=26,
            ymin=2e3,
            ymax=5e8,
            ylabel={number of pixels},
            xtick={1.5,5,8.5,13.5,18,21,24.5},
            minor xtick={3,7,10,17,19,23},
            xticklabels = {
                flat,
                construction,
                nature,
                vehicle,
                sky,
                object,
                human,
            },
            major x tick style = {opacity=0},
            minor x tick num = 1,
            xtick pos=left,
            ymajorgrids=true,
            every node near coord/.append style={
                    anchor=west,
                    rotate=90,
                    font=\footnotesize,
            }
            ]

    \addplot[bar shift=0pt,draw=road,          fill opacity=0.9,fill=road!80!white           , nodes near coords=road                 ] plot coordinates{ ( 1,     30030614  ) };
    \addplot[bar shift=0pt,draw=sidewalk,      fill opacity=0.8,fill=sidewalk!80!white       , nodes near coords=sidewalk             ] plot coordinates{ ( 2,     2945954   ) };

    \addplot[bar shift=0pt,draw=building,      fill opacity=0.8,fill=building!80!white       , nodes near coords=build.               ] plot coordinates{ ( 4,     7911269  ) };
    \addplot[bar shift=0pt,draw=fence,         fill opacity=0.8,fill=fence!80!white          , nodes near coords=fence                ] plot coordinates{ ( 5,     342958    ) };
    \addplot[bar shift=0pt,draw=wall,          fill opacity=0.8,fill=wall!80!white           , nodes near coords=wall                 ] plot coordinates{ ( 6,     135826    ) };

    \addplot[bar shift=0pt,draw=vegetation,    fill opacity=0.8,fill=vegetation!80!white     , nodes near coords=veget.               ] plot coordinates{ ( 8,    11163329  ) };
    \addplot[bar shift=0pt,draw=terrain,       fill opacity=0.8,fill=terrain!80!white        , nodes near coords=terrain              ] plot coordinates{ ( 9,    335441   ) };

    \addplot[bar shift=0pt,draw=car,           fill opacity=0.8,fill=car!80!white            , nodes near coords=car\fnn{1}           ] plot coordinates{ ( 11,    2615133   ) };
    \addplot[bar shift=0pt,draw=truck,         fill opacity=0.8,fill=truck!80!white          , nodes near coords=truck\fnn{1}         ] plot coordinates{ ( 12,    785551    ) };
    \addplot[bar shift=0pt,draw=train,         fill opacity=0.8,fill=train!80!white          , nodes near coords=train\fnn{1}         ] plot coordinates{ ( 13,    391271    ) };
    \addplot[bar shift=0pt,draw=bus,           fill opacity=0.8,fill=bus!80!white            , nodes near coords=bus\fnn{1}           ] plot coordinates{ ( 14,    301743    ) };
    \addplot[bar shift=0pt,draw=bicycle,       fill opacity=0.8,fill=bicycle!80!white        , nodes near coords=bicycle\fnn{1}       ] plot coordinates{ ( 15,    19570    ) };
    \addplot[bar shift=0pt,draw=motorcycle,    fill opacity=0.8,fill=motorcycle!80!white     , nodes near coords=motorcycle\fnn{1}    ] plot coordinates{ ( 16,    2517     ) };

    \addplot[bar shift=0pt,draw=sky,           fill opacity=0.8,fill=sky!80!white            , nodes near coords=sky                  ] plot coordinates{ ( 18,    14155896   ) };

    \addplot[bar shift=0pt,draw=pole,          fill opacity=0.8,fill=pole!80!white           , nodes near coords=pole                 ] plot coordinates{ ( 20,    861216   ) };
    \addplot[bar shift=0pt,draw=traffic sign,  fill opacity=0.8,fill=traffic sign!80!white   , nodes near coords=traffic sign         ] plot coordinates{ ( 21,    297064    ) };
    \addplot[bar shift=0pt,draw=traffic light, fill opacity=0.8,fill=traffic light!80!white  , nodes near coords=traffic light        ] plot coordinates{ ( 22,    169004    ) };

    \addplot[bar shift=0pt,draw=person,        fill opacity=0.8,fill=person!80!white         , nodes near coords=person\fnn{1}        ] plot coordinates{ ( 24,    353093   ) };
    \addplot[bar shift=0pt,draw=rider,         fill opacity=0.8,fill=rider!80!white          , nodes near coords=rider\fnn{1}         ] plot coordinates{ ( 25,    17710    ) };

    \node at (axis cs:18.5,5e8) [draw=none,anchor=north west,font=\footnotesize] {\fnn{1} instances are distinguished};

    \end{axis}
    \end{tikzpicture}
    \caption{Number of annotated pixels per class for \emph{Foggy Driving}}
    \label{fig:foggy:driving:stats:segmentation}
\end{figure*}

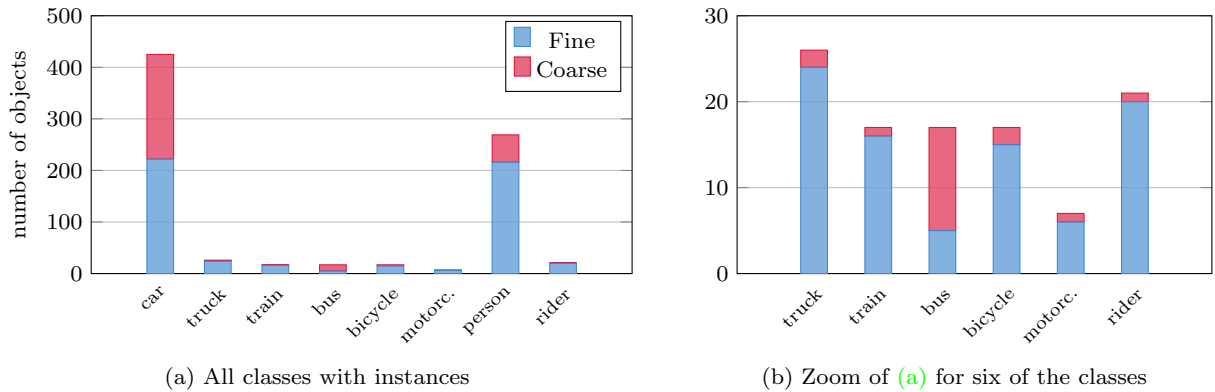
\begin{figure*}[tb]
  \centering
  \subfloat[All classes with instances]{
    \pgfplotstableread[row sep=\\]{
      1   222   203\\
      2    24     2\\
      3    16     1\\
      4     5    12\\
      5    15     2\\
      6     6     1\\
      7   216    53\\
      8    20     1\\
    }\objectsvsclass

    \begin{tikzpicture}
    \tikzstyle{every node}=[font=\small]
    \begin{axis}[
            ybar stacked,
            width=0.5\textwidth,
            height=5cm,
            xmin=-0.2,
            xmax=9.2,
            ymin=0,
            ymax=500,
            ytick={0,100,200,300,400,500},
            ymajorgrids=true,
            ylabel={number of objects},
            xtick=data,
            xticklabels = {
                car,
                truck,
                train,
                bus,
                bicycle,
                motorc.,
                person,
                rider,
            },
            xtick pos=left,
            major x tick style = {opacity=0},
            x tick label style = {rotate=45,anchor=north east,font=\scriptsize},
            ]

    \addplot[fill opacity=0.8,  draw=fine,  fill=fine!80!white] table[x index=0, y index=1] \objectsvsclass;
    \addplot[fill opacity=0.8,draw=coarse,fill=coarse!80!white] table[x index=0, y index=2] \objectsvsclass;

    \legend{Fine,Coarse}

    \end{axis}
    \end{tikzpicture}
    \label{fig:foggy:driving:stats:objects:all}
  }
  \hfil
  \subfloat[Zoom of \protect\subref{fig:foggy:driving:stats:objects:all} for six of the classes]{
    \pgfplotstableread[row sep=\\]{
      1    24     2\\
      2    16     1\\
      3     5    12\\
      4    15     2\\
      5     6     1\\
      6    20     1\\
    }\objectsvsclasssmall

    \begin{tikzpicture}
    \tikzstyle{every node}=[font=\small]
    \begin{axis}[
            ybar stacked,
            width=0.45\textwidth,
            height=5cm,
            xmin=-0.2,
            xmax=7.2,
            ymin=0,
            ymax=30,
            ymajorgrids=true,
            xtick=data,
            xticklabels = {
                truck,
                train,
                bus,
                bicycle,
                motorc.,
                rider,
            },
            xtick pos=left,
            major x tick style = {opacity=0},
            x tick label style = {rotate=45,anchor=north east,font=\scriptsize},
            ]

    \addplot[fill opacity=0.8,  draw=fine,  fill=fine!80!white] table[x index=0, y index=1] \objectsvsclasssmall;
    \addplot[fill opacity=0.8,draw=coarse,fill=coarse!80!white] table[x index=0, y index=2] \objectsvsclasssmall;

    \end{axis}
    \end{tikzpicture}
    \label{fig:foggy:driving:stats:objects:rare}
  }
  \caption{Number of objects per class in \emph{Foggy Driving}. \protect\subref{fig:foggy:driving:stats:objects:all} includes statistics for the complete set of eight classes for which instances are distinguished, whereas \protect\subref{fig:foggy:driving:stats:objects:rare} presents a zoomed version of \protect\subref{fig:foggy:driving:stats:objects:all} for six of these classes}
  \label{fig:foggy:driving:stats:objects}
\end{figure*}

In total, 33 images have been finely annotated (cf.\ the last three rows of \autoref{fig:sem:seg}) in the aforementioned procedure, and the rest 68 images have been coarsely annotated (cf.\ the top three rows of \autoref{fig:sem:seg}). We provide per-class statistics for the pixel-level semantic annotations of \emph{Foggy Driving} in \autoref{fig:foggy:driving:stats:segmentation}. Furthermore, statistics for the number of objects in the bounding box annotations are shown in \autoref{fig:foggy:driving:stats:objects}. Because of the coarse annotation that is created for one part of \emph{Foggy Driving}, we do not use this part in evaluation of object detection approaches, as difficult objects that are not included in the annotations may be detected by a good method and missed by a comparatively worse method, resulting in incorrect comparisons with respect to precision. On the contrary, the coarsely annotated images are used without such issues in evaluation of semantic segmentation approaches, since predictions at unlabeled pixels are simply ignored and thus do not affect the measured performance.

\emph{Foggy Driving} may have a smaller size than other recent datasets for semantic scene understanding, however, it features challenging foggy scenes with comparatively high complexity. As \autoref{table:foggy:driving:stats:comparison} shows, the subset of 33 images with fine annotations is roughly on par with Cityscapes regarding the average number of humans and vehicles per image. In total, \emph{Foggy Driving} contains more than 500 vehicles and almost 300 humans. We also underline the fact that \autoref{table:foggy:driving:stats:comparison} compares \emph{Foggy Driving} --- a dataset used purely for testing --- against the unions of training and validation sets of KITTI~\cite{kitti} and Cityscapes, which are much larger than their respective testing sets that would provide a better comparison.

\begin{table}[!tb]
    \centering
    \caption{Absolute and average number of annotated pixels, humans and vehicles for \emph{Foggy Driving} (``Ours''), KITTI and Cityscapes. ``h/im'' stands for humans per image and ``v/im'' for vehicles per image. Only the training and validation sets of KITTI and Cityscapes are considered}
    \label{table:foggy:driving:stats:comparison}
    \def\arraystretch{1}
    \setlength\tabcolsep{4pt}
    \begin{tabular*}{\linewidth}{l @{\extracolsep{\fill}} rrrrr}
    \toprule
    & pixels & humans & vehicles & h/im & v/im\\
    \midrule
    Ours (fine) & 38.3M & 236 & 288 & \best{7.2} & 8.7\\
    Ours (coarse) & 34.6M & 54 & 221 & 0.8 & 3.3\\
    KITTI & 0.23G & 6.1k & 30.3k & 0.8 & 4.1\\
    Cityscapes & \best{9.43G} & \best{24.0k} & \best{41.0k} & 7.0 & \best{11.8}\\
    \bottomrule
    \end{tabular*}
\end{table}

As a final note, we identify the subset of the 19 annotated classes that occur frequently in \emph{Foggy Driving}. These ``frequent'' classes either have a larger number of total annotated pixels, \eg{}\emph{road}, or a larger number of total annotated polygons or instances, \eg{}\emph{pole} and \emph{person}, compared to the rest of the classes. They are: \emph{road}, \emph{sidewalk}, \emph{building}, \emph{pole}, \emph{traffic light}, \emph{traffic sign}, \emph{vegetation}, \emph{sky}, \emph{person}, and \emph{car}. In the experiments that follow in Section~\ref{sec:results:sl:segmentation}, we occasionally use this set of frequent semantic classes as an alternative to the complete set of semantic classes for averaging per-class scores, in order to further verify results based only on classes with plenty of examples.

\section{Supervised Learning with Synthetic Fog}
\label{sec:learning:sl}

We first show that our synthetic \emph{Foggy Cityscapes-refined} dataset can be used per se for successfully adapting modern CNN models to the condition of fog with the usual supervised learning paradigm. Our experiments focus primarily on the task of semantic segmentation and additionally include comparisons on the task of object detection, evidencing clearly the usefulness of our synthetic foggy data in understanding the semantics of \emph{real} foggy scenes such as those in \emph{Foggy Driving}.

More specifically, the general outline of our main experiments can be summarized in two steps:
\begin{enumerate}
\item fine-tuning a model that has been trained on the original Cityscapes dataset for clear weather by using only synthetic images of \emph{Foggy Cityscapes-refined}, and
\item evaluating the fine-tuned model on \emph{Foggy Driving} and showing that its performance is improved compared to the original, clear-weather model. Thus, the reported results pertain to \emph{Foggy Driving} unless otherwise mentioned.
\end{enumerate}
In other words, all models are ultimately evaluated on data from a different domain than that of the data on which they have been fitted, revealing their true generalization potential on previously unseen foggy scenes.

We also consider dehazing as an optional preprocessing step before feeding the input images to semantic segmentation models for training and testing, and examine the effect of this dehazing preprocessing on the performance of such a model using state-of-the-art dehazing methods. The effect of dehazing on semantic segmentation performance is additionally correlated with its utility for human understanding of foggy scenes by conducting a user study on Amazon Mechanical Turk.

\subsection{Semantic Segmentation}
\label{sec:results:sl:segmentation}

Our model of choice for conducting experiments on semantic segmentation with the supervised pipeline is the modern dilated convolutions network (DCN)~\cite{dilated:convolution}. In particular, we make use of the publicly available \emph{Dilation10} model, which has been trained on the 2975 images of the training set of Cityscapes. We wish to note that this model was originally trained and tested on $1396\times{}1396$ image crops by the authors of~\cite{dilated:convolution}, but due to GPU memory limitations we train it on $756\times{}756$ crops and test it on $700\times{}700$ crops. Still, \emph{Dilation10} enjoys a fair mean intersection over union (IoU) score of 34.9\% on \emph{Foggy Driving}.

In the following experiments of Section~\ref{sec:results:sl:segmentation}, we fine-tune \emph{Dilation10} on the training set of \emph{Foggy Cityscapes-refined} which consists of 498 images, and reserve the 52 images of the respective validation set for additional evaluation. In particular, we fine-tune all layers of the original model for 3k iterations (ca.\ 6 epochs) using mini-batches of size 1. Unless otherwise mentioned, the attenuation coefficient $\beta$ used in \emph{Foggy Cityscapes} is equal to $0.01$.

Overall, we consider four different options with respect to dehazing preprocessing: applying no dehazing at all, dehazing with multi-scale convolutional neural networks (MSCNN)~\cite{RLZ+16}, dehazing using the dark channel prior (DCP)~\cite{dark:channel}, and non-local image dehazing~\cite{nonlocal:image:dehazing}. Unless otherwise specified, no dehazing is applied. Our experimental protocol is consistent with respect to dehazing preprocessing: the same option for dehazing preprocessing is used both at training time and test time. More specifically, at training time we first process the synthetic foggy images of \emph{Foggy Cityscapes-refined} according to the specified option for dehazing preprocessing and then use the processed images as input for fine-tuning \emph{Dilation10}. At evaluation time, we process the images in \emph{Foggy Driving} with the same dehazing preprocessing that was used at training time (if any was), and use the processed images to test the fine-tuned model.

\noindent
\textbf{Benefit of Fine-tuning on Synthetic Fog}. Our first experiment evidences the benefit of fine-tuning on \emph{Foggy Cityscapes-refined} for improving semantic segmentation performance on \emph{Foggy Driving}. \autoref{table:finetuning:dehazing} presents comparative performance of the original \emph{Dilation10} model against its fine-tuned counterparts in terms of mean IoU over all annotated classes in \emph{Foggy Driving} as well as over frequent classes only. All four options regarding dehazing preprocessing are considered. Note that we also evaluate the original \emph{Dilation10} model for all dehazing preprocessing alternatives (only relevant at test time in this case) in the first row of each part of \autoref{table:finetuning:dehazing}. Indeed, all fine-tuned models outperform \emph{Dilation10} irrespective of the type of dehazing preprocessing that is applied, both for mean IoU over all classes and over frequent classes only. The best-performing fine-tuned model, which we refer to as \emph{FT-0.01}, involves no dehazing and outperforms \emph{Dilation10} significantly, \ie{}by 3\% for mean IoU over all classes and 5\% for mean IoU over frequent classes. Note additionally that \emph{FT-0.01} has been fine-tuned on only 498 training images of \emph{Foggy Cityscapes-refined}, compared to the 2975 training images of Cityscapes for \emph{Dilation10}.

\begin{table}[!tb]
  \centering
  \caption{Performance comparison on \emph{Foggy Driving} of \emph{Dilation10} versus fine-tuned versions of it using \emph{Foggy Cityscapes-refined}, for four options regarding dehazing preprocessing. ``FT'' stands for using fine-tuning and ``W/o FT'' for not using fine-tuning}
  \label{table:finetuning:dehazing}
  Mean IoU over \emph{all} classes (\%)
  \setlength\tabcolsep{4pt}
  \begin{tabular*}{\linewidth}{l @{\extracolsep{\fill}} cccc}
  \toprule
  & No dehazing & MSCNN & DCP & Non-local\\
  \midrule
  W/o FT & 34.9 & 34.7 & 29.9 & 29.3\\
  FT & \best{37.8} & \best{37.1} & \best{37.4} & \best{36.6}\\
  \bottomrule
  \end{tabular*}
  \\[\baselineskip]
  Mean IoU over \emph{frequent} classes in \emph{Foggy Driving} (\%)
  \setlength\tabcolsep{4pt}
  \begin{tabular*}{\linewidth}{l @{\extracolsep{\fill}} cccc}
  \toprule
  & No dehazing & MSCNN & DCP & Non-local\\
  \midrule
  W/o FT & 52.4 & 52.4 & 45.5 & 46.2\\
  FT & \best{57.4} & \best{56.2} & \best{56.7} & \best{55.1}\\
  \bottomrule
  \end{tabular*}
\end{table}

\noindent
\textbf{Comparison of Fog Simulation Approaches}. Next, we compare in \autoref{table:fog:simulation:lr:policy} the utility of our proposed fog simulation method for generating useful synthetic training data in terms of semantic segmentation performance on \emph{Foggy Driving}, against two alternative approaches: the baseline that we considered in \autoref{fig:fog:simulation} and a truncated version of our method, where we omit the guided filtering step. We consider two different policies for the learning rate when fine-tuning on \emph{Foggy Cityscapes-refined}: a constant learning rate of $10^{-5}$ and a polynomially decaying learning rate, commonly referred to as ``poly''~\cite{DeepLab}, with a base learning rate of $10^{-5}$ and a power parameter of $0.9$. Our method for fog simulation consistently outperforms the two baselines and the ``poly'' learning rate policy allows the model to be fine-tuned more effectively than the constant policy. In all other experiments with DCN, we use the ``poly'' learning rate policy with the parameters specified above for fine-tuning.

\begin{table}[!tb]
  \centering
  \caption{Performance comparison on \emph{Foggy Driving} of various fine-tuned versions of \emph{Dilation10} that correspond to different fog simulation methods for generating the training dataset \emph{Foggy Cityscapes-refined} that is used for fine-tuning, and different learning rate policies during fine-tuning. Mean IoU (\%) over \emph{all} classes is used to report results}
  \label{table:fog:simulation:lr:policy}
  \setlength\tabcolsep{4pt}
  \begin{tabular*}{\linewidth}{l @{\extracolsep{\fill}} cc}
  \toprule
  & Constant l.r. & ``Poly'' l.r.\\
  \midrule
  Nearest neighbor & 32.9 & 36.2 \\
  Ours w/o guided filtering & 33.0 & 36.8 \\
  Ours & \best{34.4} & \best{37.8} \\
  \bottomrule
  \end{tabular*}
\end{table}

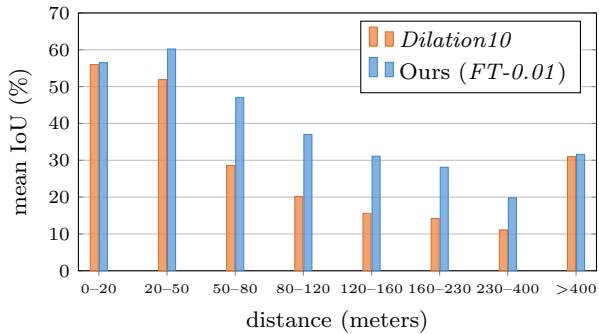
\begin{figure}[tb]
    \centering
    \pgfplotstableread{
    0   56.0   56.1   56.5
    1   51.9   56.0   60.2
    2   28.6   39.2   47.0
    3   20.2   25.8   37.0
    4   15.6   20.4   31.1
    5   14.2   19.8   28.1
    6   11.1   14.8   19.8
    7   31.0   25.9   31.6
    }\iouvsdistance

    \begin{tikzpicture}
    \tikzstyle{every node}=[font=\small]
    \begin{axis}[
            ybar=\pgflinewidth,
            width=\columnwidth,
            height=5cm,
            xmin=-0.3,
            xmax=7.3,
            ymin=0,
            ymax=70,
            ymajorgrids=true,
            bar width=3pt,
            ylabel={mean IoU (\%)},
            xlabel={distance (meters)},
            xtick=data,
            xticklabels = {
                0--20 ,
                20--50 ,
                50--80 ,
                80--120 ,
                120--160 ,
                160--230 ,
                230--400 ,
               \textgreater{}400 ,
            },
            ytick={0,10,20,30,40,50,60,70},
            yticklabels = {
                0,
                10,
                20,
                30,
                40,
                50,
                60,
                70,
            },
            x tick label style={font=\tiny},
            xtick pos=left,
            y tick label style={font=\scriptsize},
            major tick length=0.5ex,
            legend cell align=left,
            legend pos=north east
            ]
    \addplot[fill opacity=0.8,draw=dcn,fill=dcn!80!white] table[x index=0,y index=1] \iouvsdistance;
    \addplot[fill opacity=0.8,draw=dcn-ft-001,fill=dcn-ft-001!80!white] table[x index=0,y index=3] \iouvsdistance;

    \legend{\emph{Dilation10},Ours (\emph{FT-0.01})}

    \end{axis}
    \end{tikzpicture}
\caption{Performance of semantic segmentation models on \emph{Foggy Cityscapes-refined} at distinct ranges of scene distance from the camera}
\label{fig:distance}
\end{figure}

\noindent
\textbf{Increasing Returns at Larger Distance}. As can easily be deduced from \eqref{eq:transmission}, fog has a growing effect on the appearance of the scene as distance from the camera increases. Ideally, a model that is dedicated to foggy scenes must deliver a greater benefit for distant parts of the scene. In order to examine this aspect of semantic segmentation of foggy scenes, we use the completed, dense distance maps of Cityscapes images that have been computed as an intermediate output of our fog simulation, given that \emph{Foggy Driving} does not include depth information. In more detail, we consider the validation set of \emph{Foggy Cityscapes-refined}, the images of which are unseen both for \emph{Dilation10} and our fine-tuned models, and bin the pixels according to their value in the corresponding distance map. Each distance range is considered separately for evaluation by ignoring all pixels that do not belong to it. In \autoref{fig:distance}, we compare mean IoU of \emph{Dilation10} and \emph{FT-0.01} individually for each distance range. \emph{FT-0.01} brings a consistent gain in performance across all distance ranges. What is more, this gain is larger in both absolute and relative terms for pixels that are more than 50m away from the camera, implying that our model is able to handle better the most challenging parts of a foggy scene. Note that most pixels in the very last distance range (more than 400m away from the camera) belong to the \emph{sky} class and their appearance does not change much between the clear and the synthetic foggy images.

\begin{figure}[tb]
    \centering
    \pgfplotstableread{
    0   61.4   61.9   58.9   56.4   55.8   49.9
    1   57.5   62.1   60.7   58.5   57.5   52.2
    2   51.0   59.0   59.8   58.2   57.4   52.7
    3   40.3   52.6   55.7   56.1   55.6   52.4
    4   32.8   45.2   51.0   53.5   54.6   51.9
    5   20.0   30.7   36.9   43.4   47.0   48.6
    }\ioucrosseval

    \begin{tikzpicture}
    \tikzstyle{every node}=[font=\small]
    \begin{axis}[
        width=\columnwidth,
        xmin=0,
        xmax=5,
        ymin=13,
        ymax=65,
        axis y discontinuity=crunch,
        ymajorgrids=true,
        ylabel={mean IoU (\%)},
        xlabel={attenuation coefficient $\beta$ ($\text{m}^{-1}$)},
        xtick=data,
        xticklabels = {0,0.005,0.01,0.02,0.03,0.06},
        ytick = {20,30,40,50,60,65},
        yticklabels = {20,30,40,50,60,65,},
        x tick label style={font=\scriptsize},
        xtick pos=left,
        y tick label style={font=\scriptsize},
        major tick length=0.5ex,
        legend cell align=left,
        legend pos=south west,
    ]
    
    \addplot[
        color=dcn,
        mark=+,
        mark size=2.5pt,
        mark options={ultra thick},
        thick,
    ]
    table[
        x index=0,
        y index=1,
    ] \ioucrosseval;
    
    \addplot[
        color=dcn-ft-0005,
        mark=*,
        thick,
    ]
    table[
        x index=0,
        y index=2,
    ] \ioucrosseval;
    
    \addplot[
        color=dcn-ft-001,
        mark=diamond*,
        thick,
    ]
    table[
        x index=0,
        y index=3,
    ] \ioucrosseval;

    \addplot[
        color=dcn-ft-002,
        mark=square*,
        thick,
    ]
    table[
        x index=0,
        y index=4,
    ] \ioucrosseval;

    \addplot[
        color=dcn-ft-003,
        mark=triangle*,
        thick,
    ]
    table[
        x index=0,
        y index=5,
    ] \ioucrosseval;

    \addplot[
        color=dcn-ft-006,
        mark=x,
        thick,
    ]
    table[
        x index=0,
        y index=6,
    ] \ioucrosseval;

    \legend{\emph{Dilation10}, \emph{FT-0.005}, \emph{FT-0.01}, \emph{FT-0.02}, \emph{FT-0.03}, \emph{FT-0.06}}

    \end{axis}
    \end{tikzpicture}
    
\caption{Performance of semantic segmentation models on various versions of the validation set of \emph{Foggy Cityscapes-refined} corresponding to different values of attenuation coefficient $\beta$}
\label{fig:iou:crosseval}
\end{figure}
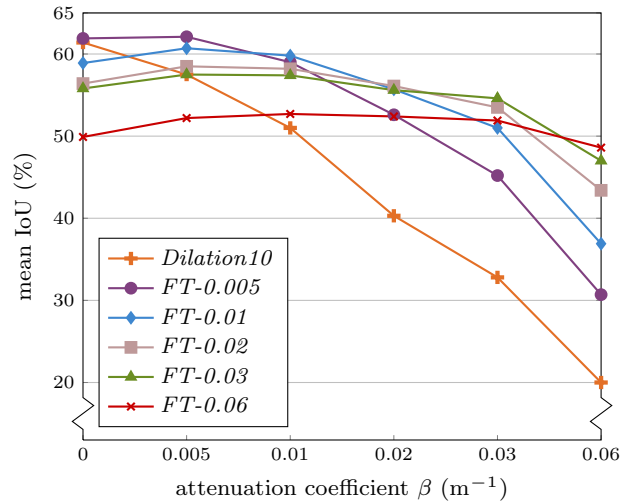

\noindent
\textbf{Generalization in Synthetic Fog across Densities}. In order to verify the ability of a model that has been fine-tuned on \emph{Foggy Cityscapes-refined} for a fixed value $\beta^{(t)}$ of the attenuation coefficient, hence fixed fog density, to generalize well to new, unseen fog densities, we evaluate the model on multiple versions of the validation set of \emph{Foggy Cityscapes-refined}, each rendered using a different value for $\beta$ which is in general not equal to $\beta^{(t)}$. In particular, we use the five different versions of \emph{Foggy Cityscapes-refined} as described in Section~\ref{sec:datasets:foggy:cityscapes} and obtain five models by fine-tuning \emph{Dilation10} on the training set of each version. In congruence with notation in previous experiments, we denote such a fine-tuned model by \emph{FT-}$\beta^{(t)}$, \eg{}\emph{FT-0.02}. Afterwards, we evaluate each of these models plus \emph{Dilation10} on the validation set of each of the five foggy versions plus the original, clear-weather version where $\beta = 0$. The mean IoU performance of the six models is presented in \autoref{fig:iou:crosseval}.
Whereas the performance of \emph{Dilation10} drops rapidly as $\beta$ increases, all five fine-tuned ``foggy'' models are more robust to changes in $\beta$ across the examined range. Analyzing the performance of each fine-tuned model individually, we observe that performance is high and fairly stable in the range $[0,\,\beta^{(t)}]$ and drops for $\beta > \beta^{(t)}$. This implies that a ``foggy'' model is able to generalize well to lighter synthetic fog than what was used to fine-tune it. Moreover, all ``foggy'' models compare favorably to \emph{Dilation10} across the largest part of the range of $\beta$, with most ``foggy'' models being beaten by \emph{Dilation10} only for clear weather. Note also that the performance gain with ``foggy'' models under foggy conditions is much larger than the corresponding performance loss for clear weather.

\noindent
\textbf{Effect of Synthetic Fog Density on Real-world Performance}. Our final experiment on semantic segmentation serves two purposes: to examine the effect of varying the fog density of the synthetic training data as well as that of dehazing preprocessing on the performance of the fine-tuned model on real foggy data. To this end, we use three of the versions of \emph{Foggy Cityscapes-refined} corresponding to the values $\{0.005,\,0.01,\,0.02\}$ for $\beta$ and consider all four options regarding dehazing preprocessing for fine-tuning \emph{Dilation10}. The performance of the 12 resulting fine-tuned models on \emph{Foggy Driving} in terms of mean IoU over all annotated classes as well as over frequent classes only is reported in \autoref{table:beta:dehazing}. We first discuss the effect of varying fog density for each dehazing option individually and defer a general comparison of the various dehazing preprocessing options to the next paragraph.

\begin{table}[!tb]
  \centering
  \caption{Performance comparison on \emph{Foggy Driving} of fine-tuned versions of \emph{Dilation10} using \emph{Foggy Cityscapes-refined}, for three different values of attenuation coefficient $\beta$ in fog simulation and four options regarding dehazing preprocessing}
  \label{table:beta:dehazing}
  Mean IoU over \emph{all} classes (\%)
  \setlength\tabcolsep{4pt}
  \begin{tabular*}{\linewidth}{l @{\extracolsep{\fill}} ccc}
  \toprule
  & $\beta=0.005$ & $\beta=0.01$ & $\beta=0.02$\\
  \midrule
  No dehazing & 37.6 & \best{37.8} & 36.1\\
  MSCNN & \best{38.3} & 37.1 & 36.9\\
  DCP & 36.6 & 37.4 & 36.1\\
  Non-local & 36.2 & 36.6 & 35.3\\
  \bottomrule
  \end{tabular*}
  \\[\baselineskip]
  Mean IoU over \emph{frequent} classes in \emph{Foggy Driving} (\%)
  \setlength\tabcolsep{4pt}
  \begin{tabular*}{\linewidth}{l @{\extracolsep{\fill}} ccc}
  \toprule
  & $\beta=0.005$ & $\beta=0.01$ & $\beta=0.02$\\
  \midrule
  No dehazing & 57.0 & \best{57.4} & 56.2\\
  MSCNN & \best{57.3} & 56.2 & 56.3\\
  DCP & 56.0 & 56.7 & 55.2\\
  Non-local & 55.1 & 55.1 & 54.5\\
  \bottomrule
  \end{tabular*}
\end{table}

The two conditions that must be met in order for the examined models to achieve better performance are:
\begin{enumerate}
\item a good matching of the distributions of the synthetic training data and the real, testing data, and \label{point:matching}
\item a clear appearance of both sets of data, in the sense that the segmentation model should have an easy job in mining discriminative features from the data. \label{point:clear}
\end{enumerate}
Focusing on the case that does not involve dehazing, we observe that the models with $\beta=0.005$ and $\beta=0.01$ perform significantly better than that with $\beta=0.02$, implying that according to point~\ref{point:matching} \emph{Foggy Driving} is dominated by scenes with light or medium fog. On the other hand, each of the three dehazing methods that are used for preprocessing has its own particularities in enhancing the appearance and contrast of foggy scenes while also introducing artifacts to the output. More specifically, MSCNN is slightly conservative in removing fog, as was found for other learning-based dehazing methods in~\cite{dehazing:survey:benchmarking}, and operates best under lighter fog, providing a significant improvement in this setting with regard to point~\ref{point:clear}. In conjunction with the light-fog character of \emph{Foggy Driving}, this explains why fine-tuning on light fog ($\beta=0.005$) combined with MSCNN preprocessing delivers one of the two best overall results. By contrast, the more aggressive DCP is known to operate better at high levels of fog, as its estimated transmission is biased towards lower values~\cite{TYW14}. The performance of models with DCP preprocessing thus peaks at medium rather than low simulated-fog density, which signifies a trade-off between removing fog to the proper extent and minimal introduction of artifacts. Non-local dehazing has also been found to operate best at medium levels of fog~\cite{dehazing:survey:benchmarking}, which results in a similar performance trend to DCP.

\noindent
\textbf{Effect of Dehazing Preprocessing on Real-world Performance and Discussion}. Comparing the four options regarding dehazing preprocessing via \autoref{table:beta:dehazing}, we observe that  applying no dehazing is the best or second best option for both measures and across all three values of $\beta$. Only MSCNN marginally beats the no-dehazing option in some cases, while overall these two options are roughly on a par. The absence of a significant performance gain on \emph{Foggy Driving} when performing dehazing preprocessing can be ascribed to generic as well as method-specific reasons.

First, in the real-world setting of \emph{Foggy Driving}, the homogeneity and uniformity assumptions of the optical model~\eqref{eq:haze:model} that is used by all examined dehazing methods may not hold exactly. Of course, this model is also used in our fog simulation, however, foggy image synthesis is a \emph{forward} problem, whereas image defogging/dehazing is an \emph{inverse} problem, hence inherently more difficult. Thus, the artifacts that are introduced by our fog simulation are likely to be less prominent than those introduced by dehazing. This fact appears to outweigh the potential increase in visibility for dehazed images as far as point~\ref{point:clear} above is concerned. An interesting insight that follows is the use of forward techniques to generate training data for hard target domains based on data from the source domain as an alternative to the application of inverse techniques to transform such target domains into the easier source domain.

Second, the optical model~\eqref{eq:haze:model}, on which most of the popular dehazing approaches rely, assumes a \emph{linear} relation between the irradiance at a pixel and the actual value of the pixel in the processed hazy image. Therefore, these approaches require that an initial gamma correction step be applied before dehazing, otherwise their performance may deteriorate significantly. This in turn implies that the value of gamma must be known for each image, which is \emph{not} the case for Cityscapes and \emph{Foggy Driving}. Manually searching for ``best'' per-image values is also infeasible for these large datasets. In the absence of any further information, we have used a constant value of 1 for gamma as the authors of~\cite{nonlocal:image:dehazing} recommend, which is probably suboptimal for most of the images. We thus wish to point out that future work on outdoor datasets, whether considering fog/haze or not, should ideally record the value of gamma for each image, so that dehazing methods can show their full potential on such datasets.

Specifically for DCP, performance decreases compared to MSCNN partly due to the light-fog character of \emph{Foggy Driving} which does not match the optimal operating point of DCP. On the other hand, non-local dehazing uses a different model for estimating atmospheric light than the one that is shared by our fog simulation, MSCNN, and DCP, and thus already faces greater difficulty in dehazing images from \emph{Foggy Cityscapes}. 

\subsection{Linking the Objective and Subjective Utility of Dehazing Preprocessing in Foggy Scene Understanding}
\label{sec:dehazing:helpful}

\begin{figure*}
\begin{tabular}{cccccc}
\hspace{-2mm}
\includegraphics[width=.24\linewidth, height=24mm]{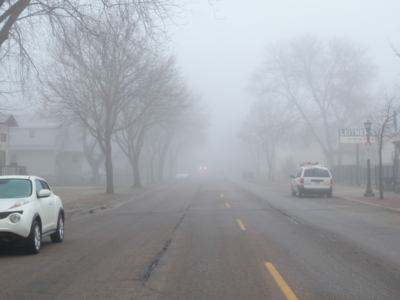} & 
\hspace{-4mm} 
\includegraphics[width=.24\linewidth, height=24mm]{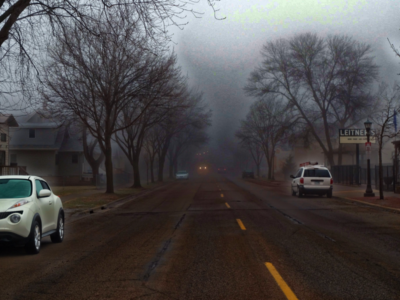} &
\hspace{-4mm} 
\includegraphics[width=.24\linewidth, height=24mm]{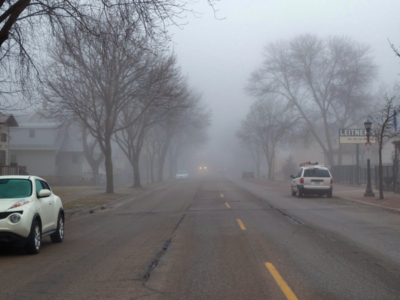} &
\hspace{-4mm} 
\includegraphics[width=.24\linewidth, height=24mm]{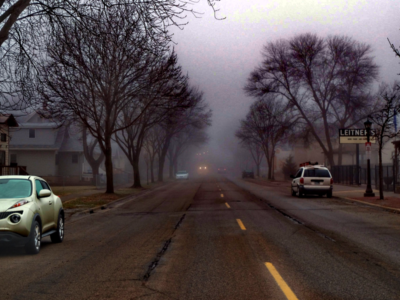} \\ 
\hspace{-2mm}
\includegraphics[width=.24\linewidth, height=24mm]{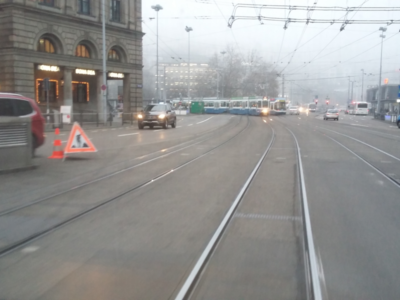} & 
\hspace{-4mm} 
\includegraphics[width=.24\linewidth, height=24mm]{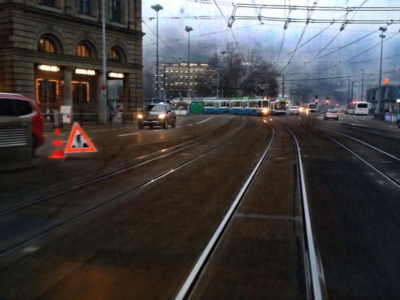} &
\hspace{-4mm} 
\includegraphics[width=.24\linewidth, height=24mm]{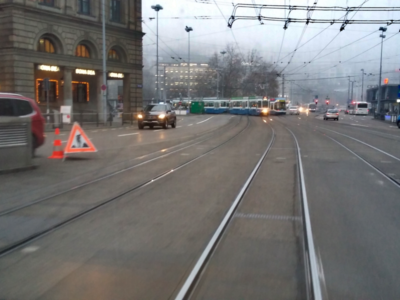} &
\hspace{-4mm} 
\includegraphics[width=.24\linewidth, height=24mm]{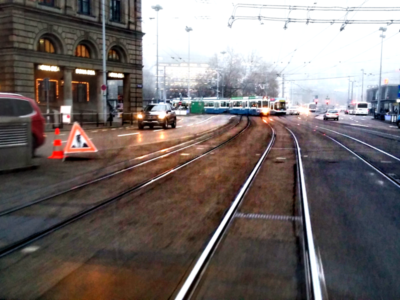} \\ 
\text{(a) Foggy}  & \text{(b) DCP}  & \text{(c) MSCNN}  & \text{(d) Non-local} \\
\end{tabular}
\caption{Example images from \emph{Foggy Driving} and their dehazed versions using three state-of-the-art dehazing methods that are examined in our experiments}
\label{fig:dehazing:example}
\end{figure*}

\begin{figure}
  \begin{center}
\includegraphics[width=.95\linewidth]{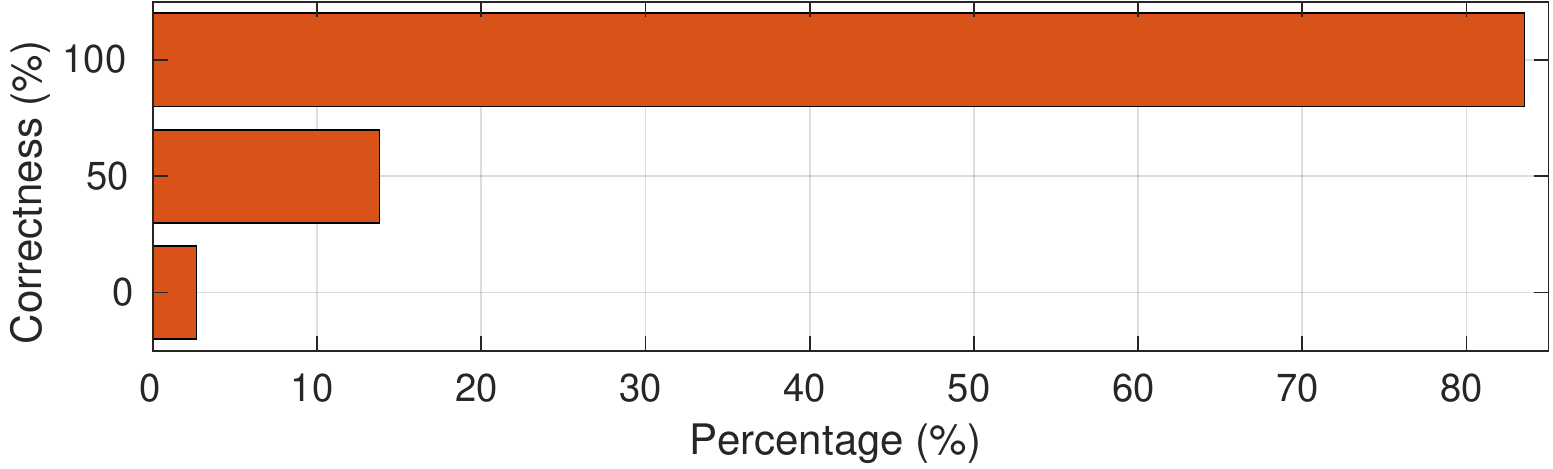} 
  \end{center}
  \vspace{-2mm}
    \caption{Quality of our user survey on AMT, computed using known-answer questions}
\label{fig:quality:usersurvey}
\end{figure}

\begin{figure*}
\begin{tabular}{cccccc}
\begin{turn}{90} \quad \small{\text{Subjective}} \end{turn} &  \includegraphics[width=.22\linewidth, height=20mm]{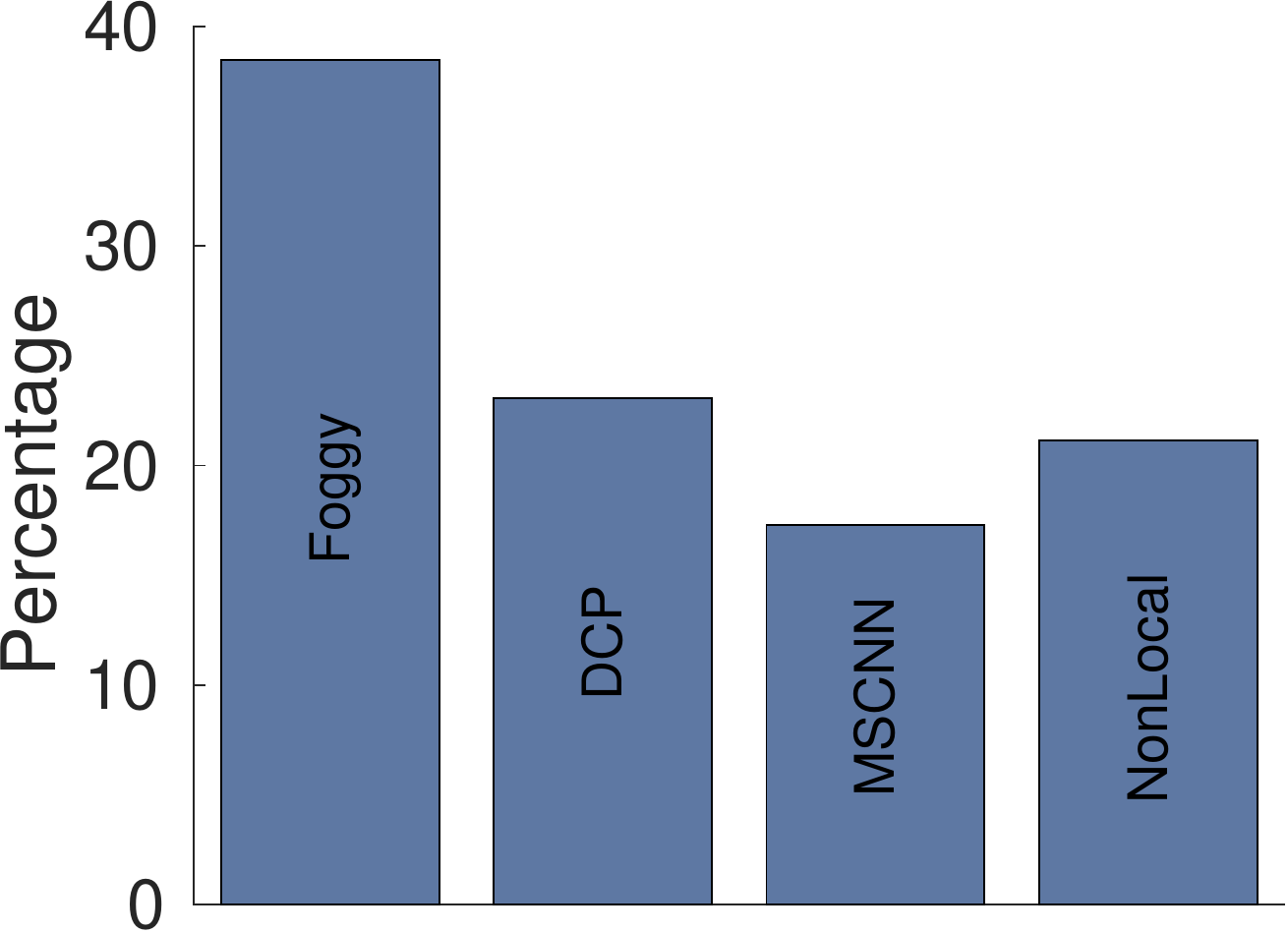} &
\hspace{-4mm} 
\includegraphics[width=.22\linewidth, height=20mm]{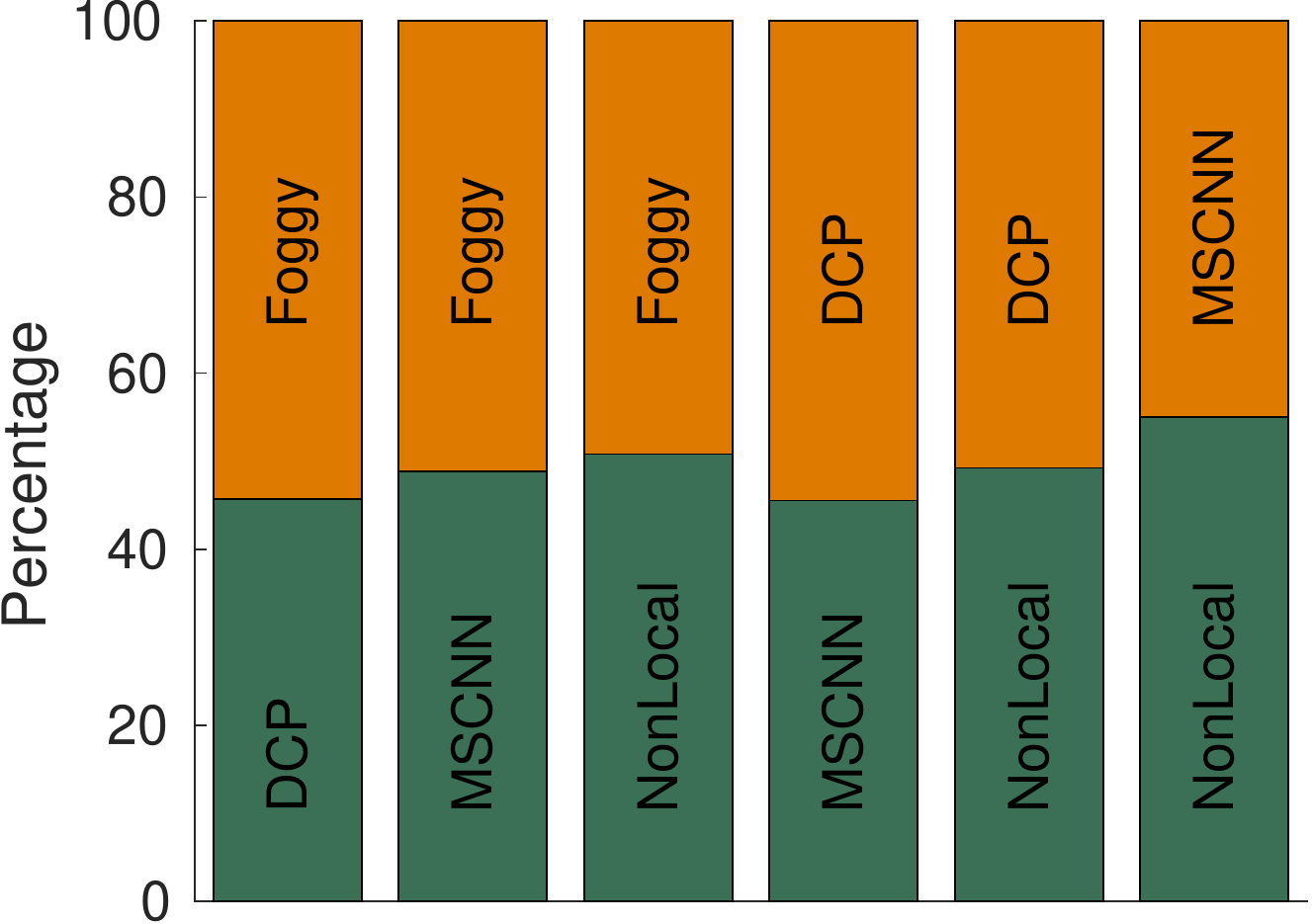} &
\hspace{4mm} 
\includegraphics[width=.22\linewidth, height=20mm]{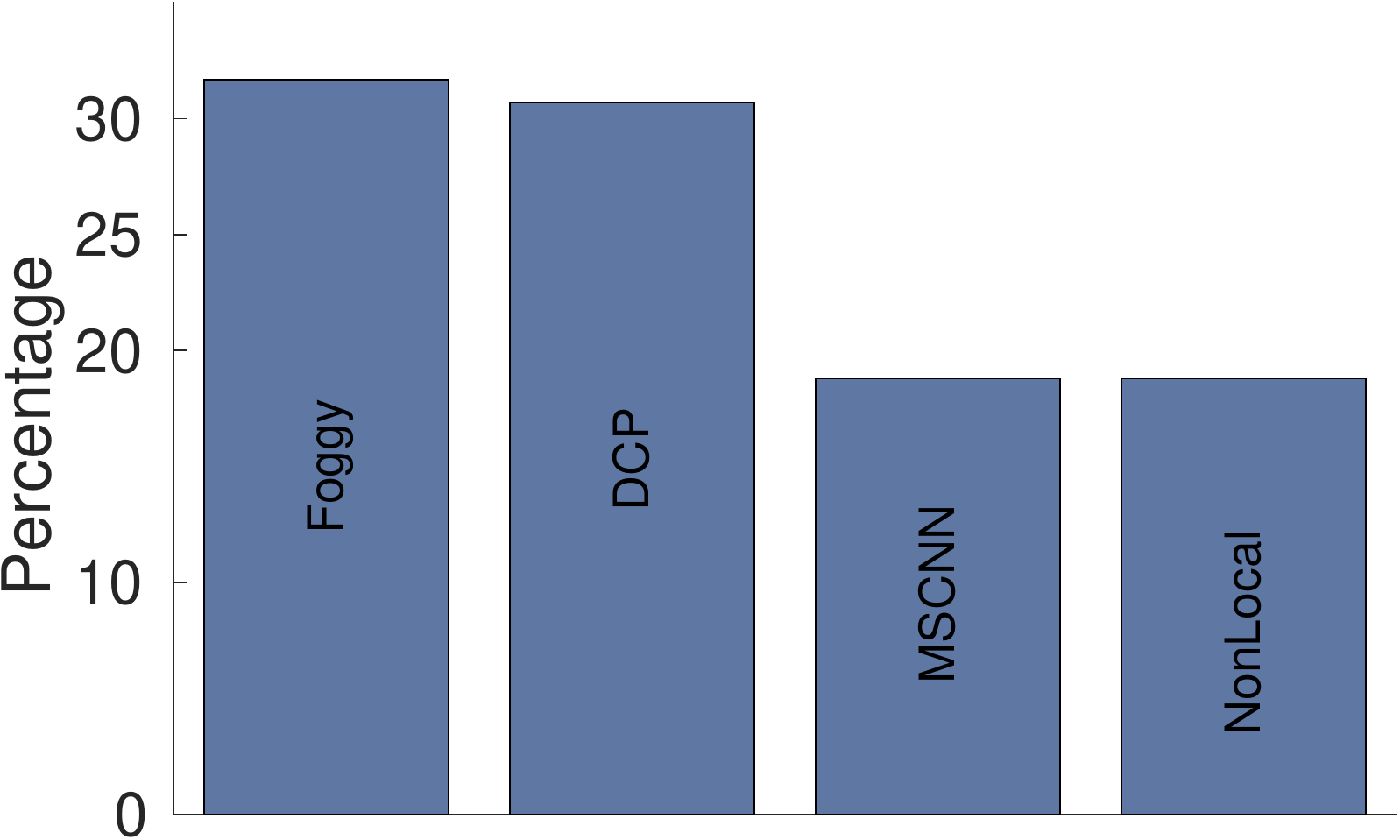} &
\hspace{-4mm} 
\includegraphics[width=.22\linewidth, height=20mm]{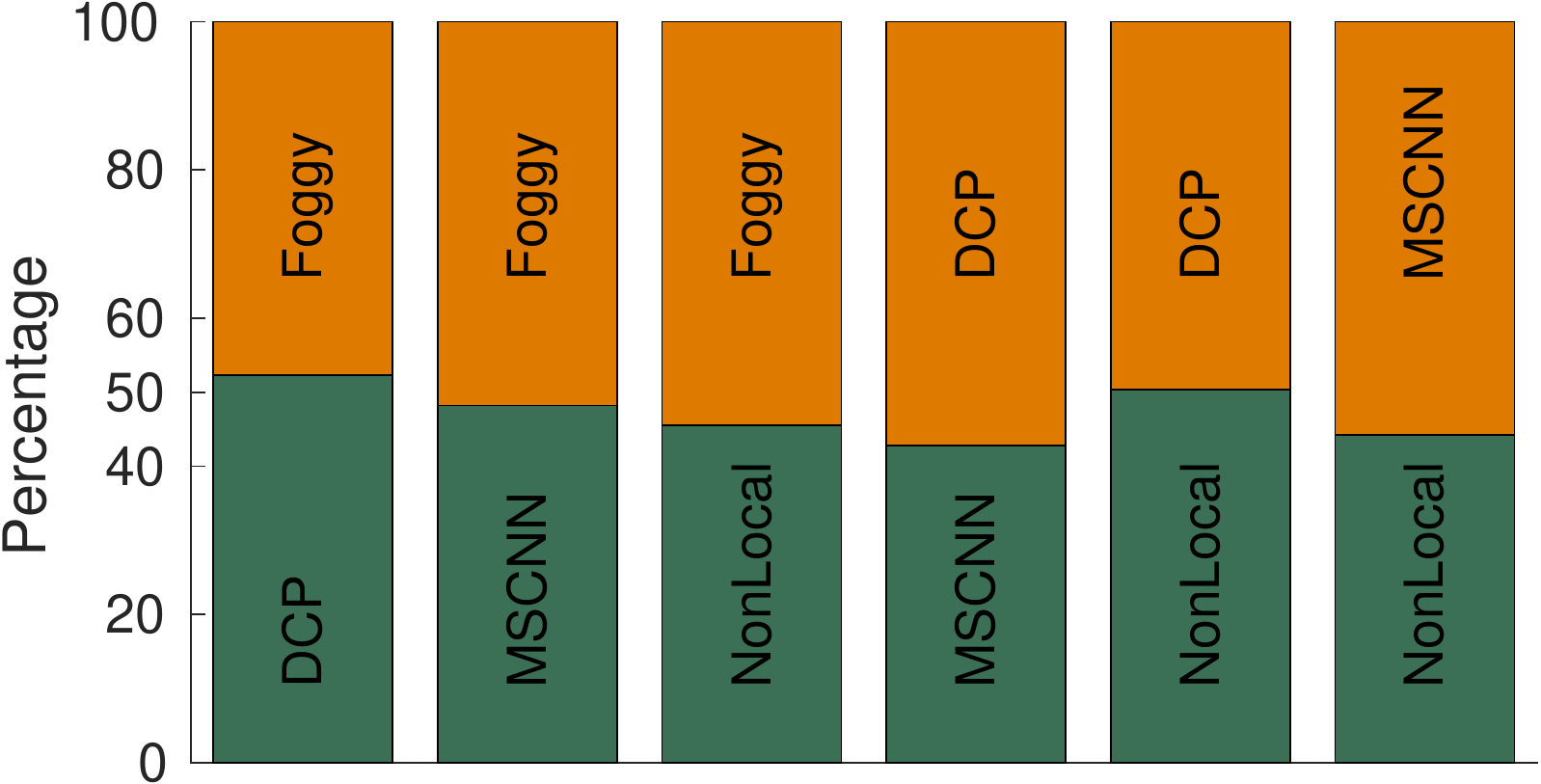} \\ 
\begin{turn}{90} \quad \small{\text{Objective}} \end{turn} &
\includegraphics[width=.22\linewidth, height=20mm]{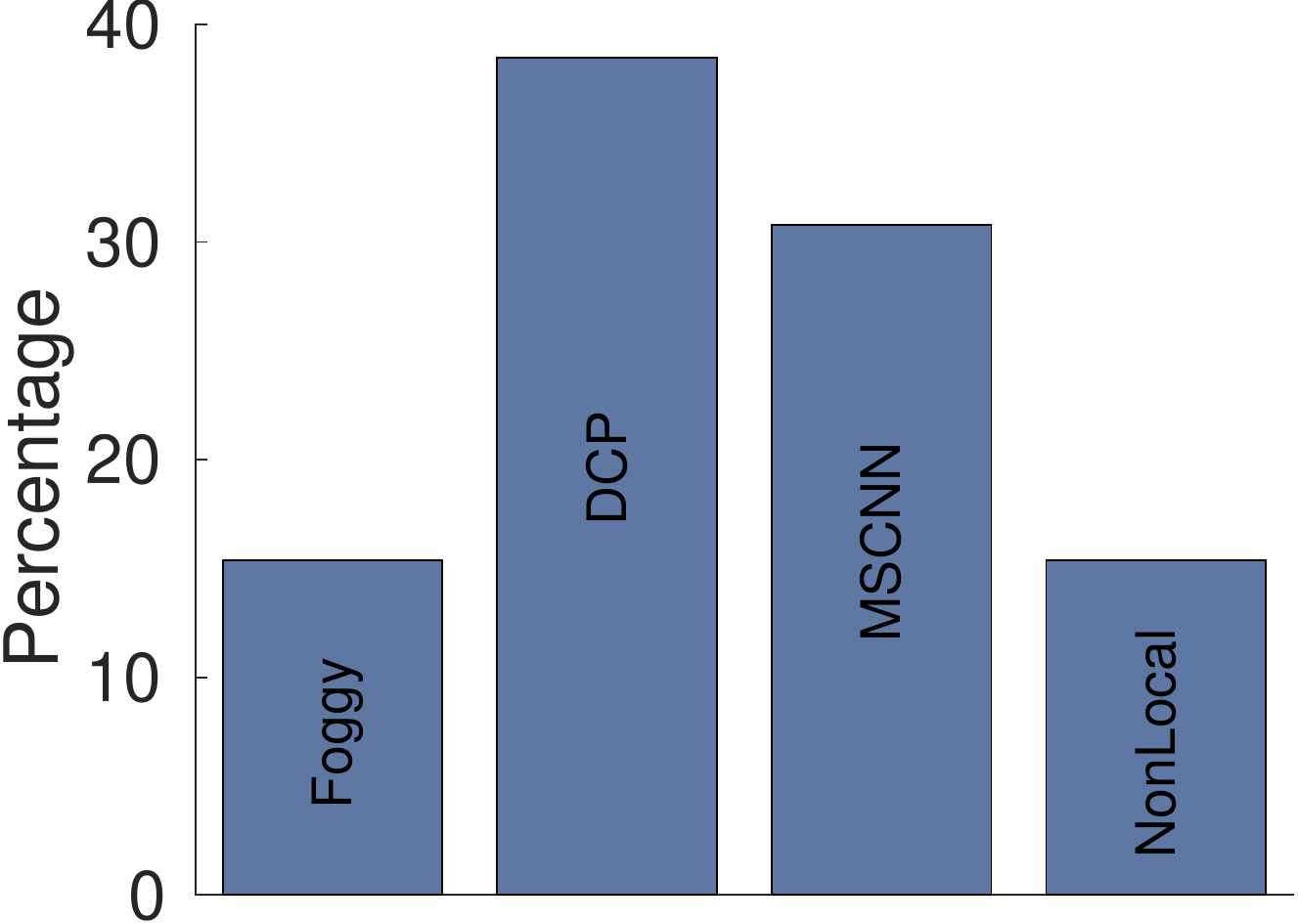} & 
\hspace{-4mm} 
\includegraphics[width=.22\linewidth, height=20mm]{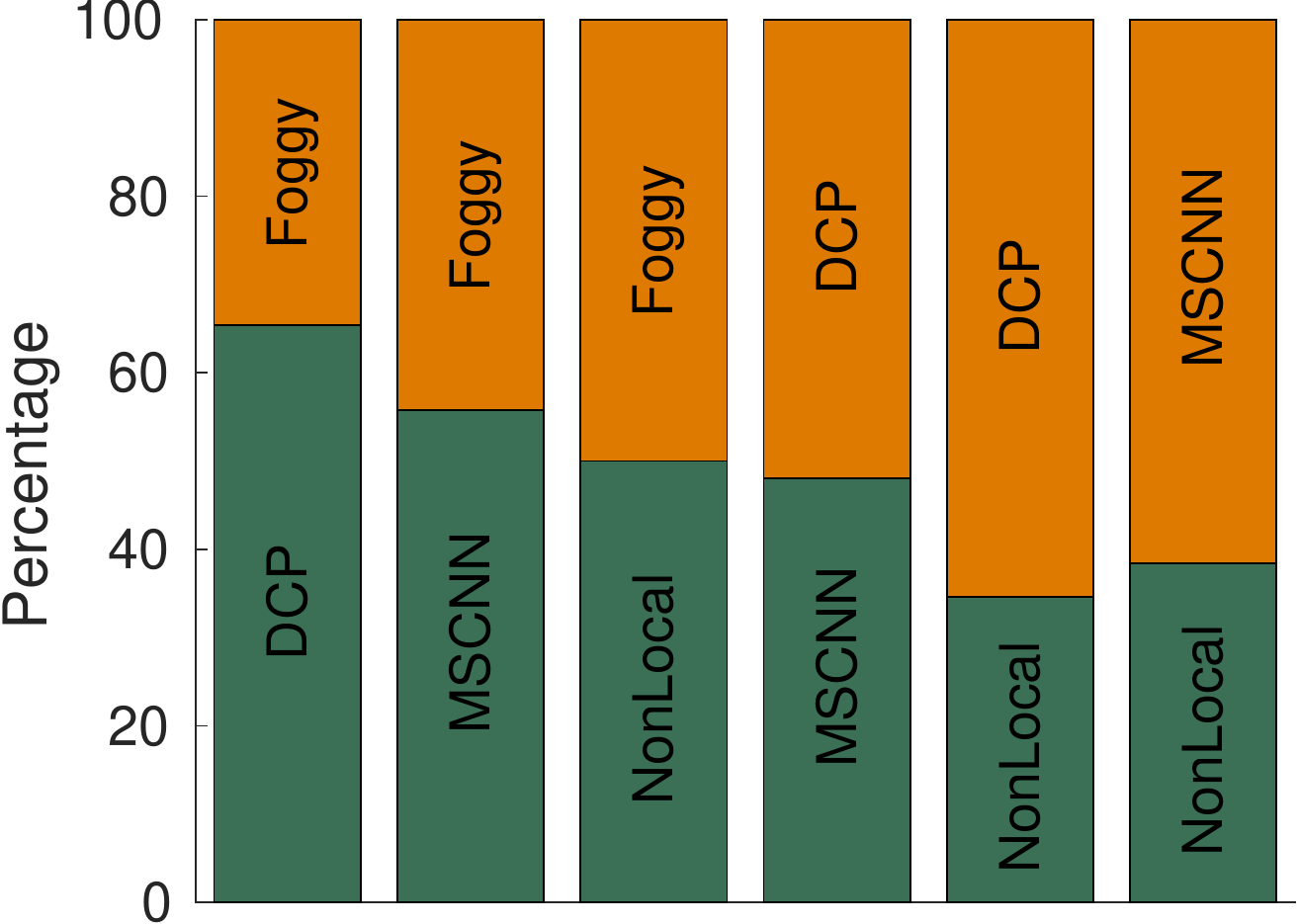} &
\hspace{4mm} 
\includegraphics[width=.22\linewidth, height=20mm]{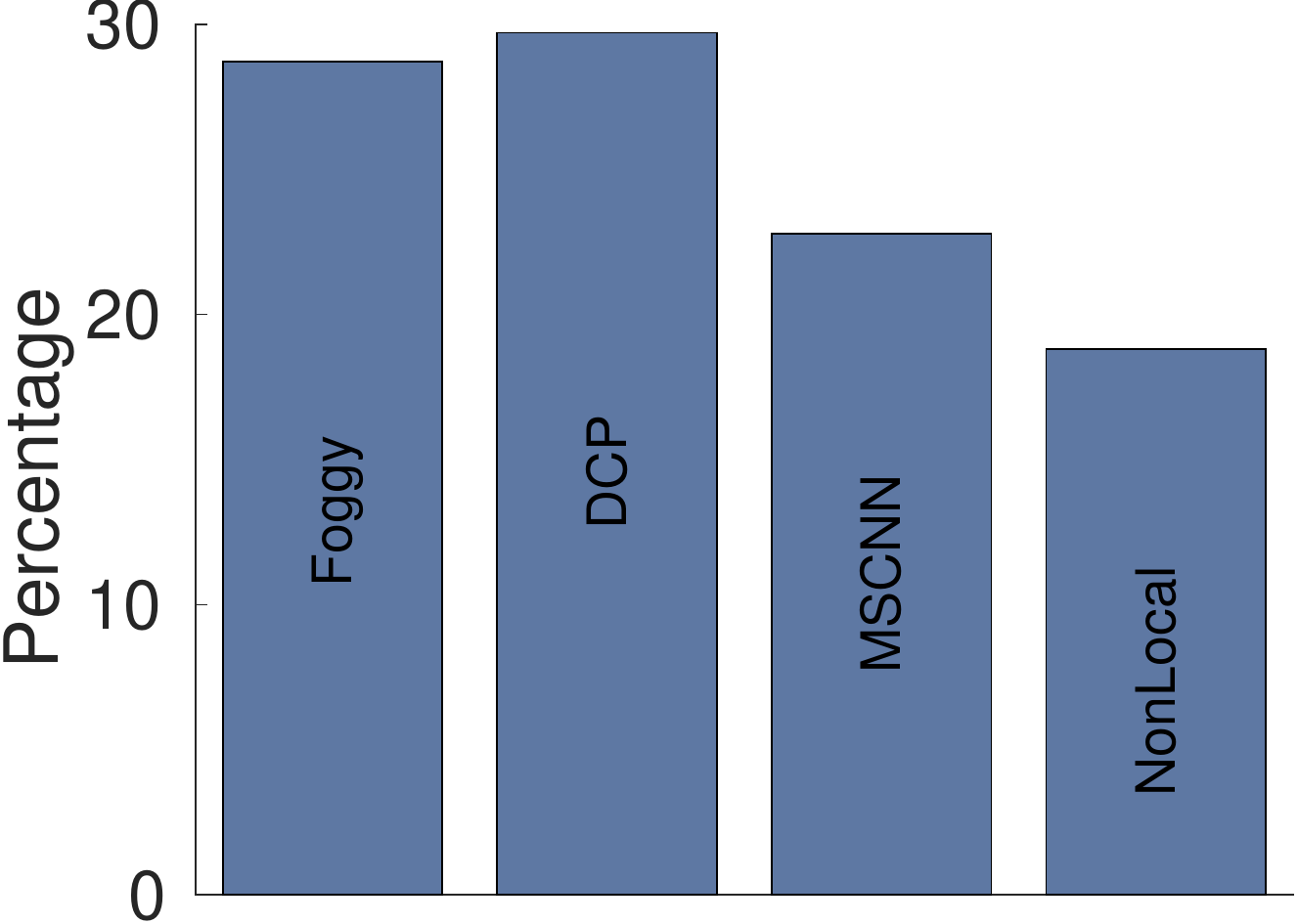} &
\hspace{-4mm} 
\includegraphics[width=.22\linewidth, height=20mm]{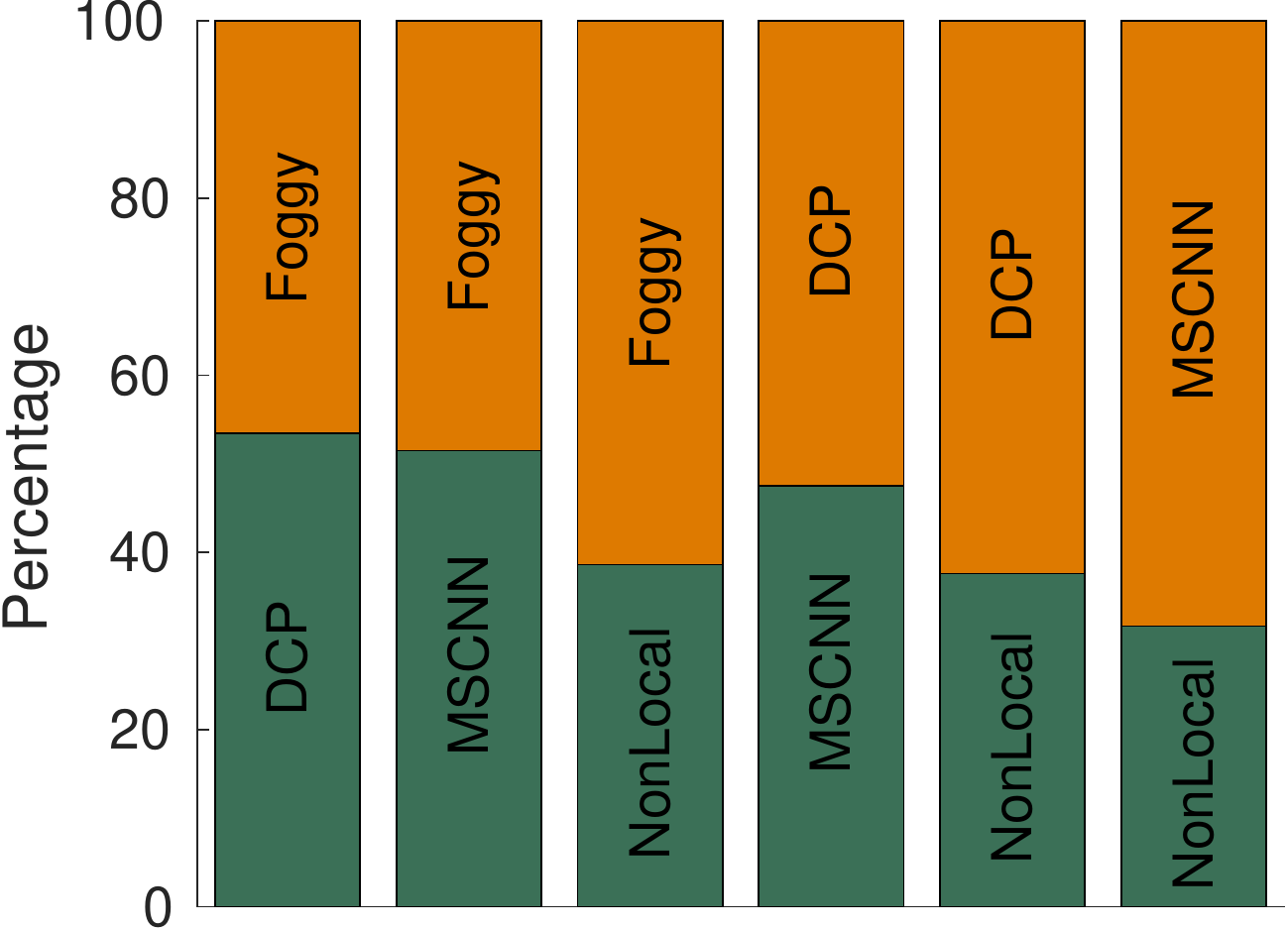} \\ 

\begin{turn}{90} \quad \quad \small{\text{Correlation}} \end{turn} &
\multicolumn{2}{c}{\includegraphics[width=.35\linewidth, height=25mm]{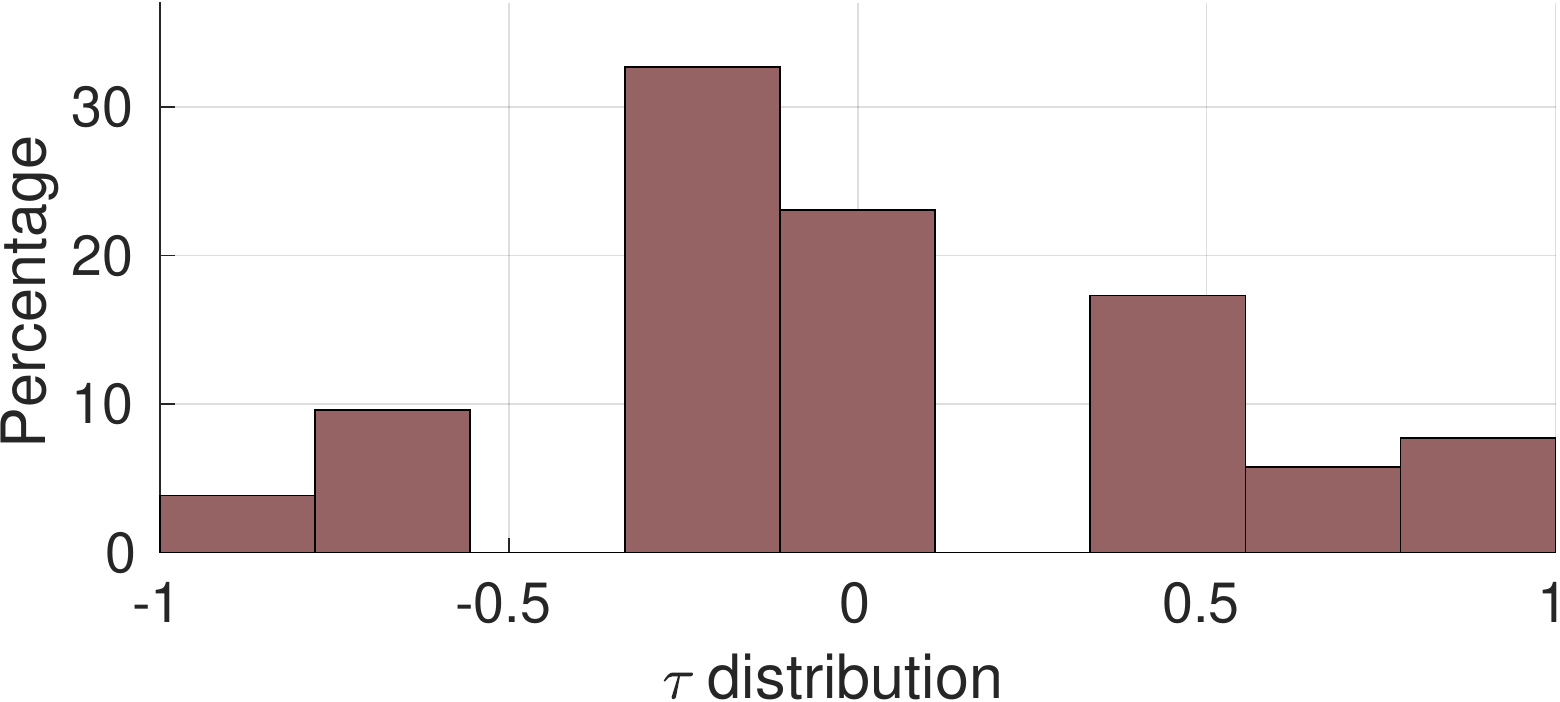}} &  \multicolumn{2}{c}{\includegraphics[width=.35\linewidth, height=25mm]{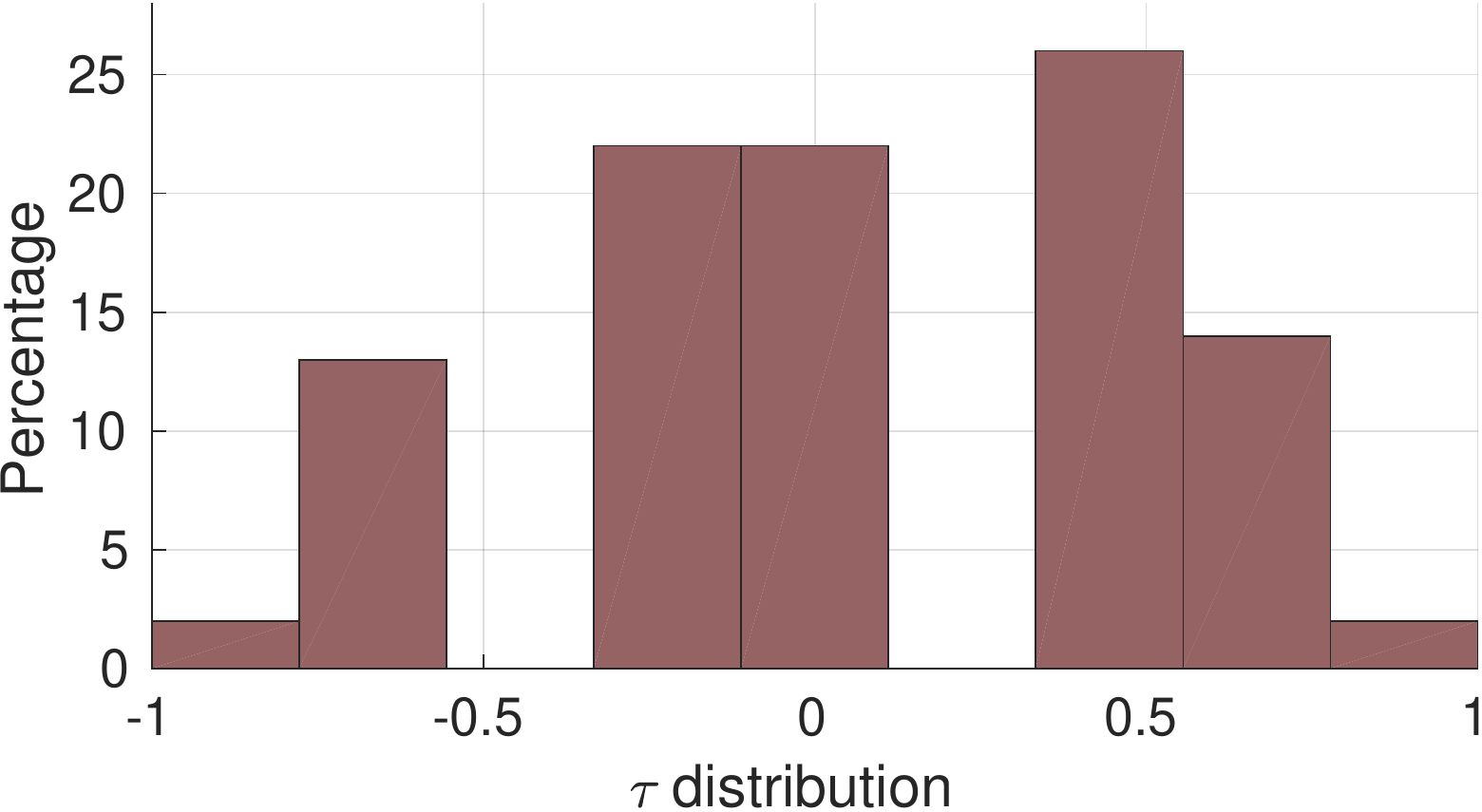}} \\
& \multicolumn{2}{c}{\text{(a) \emph{Foggy Cityscapes-refined}}} & \multicolumn{2}{c}{\text{(b) \emph{Foggy Driving}}} \\
\end{tabular}
\caption{Comparison of four options for dehazing preprocessing, \ie{}no dehazing (``Foggy''), ``DCP''~\cite{dark:channel}, ``MSCNN''~\cite{RLZ+16}, and ``NonLocal''~\cite{nonlocal:image:dehazing}, on (a) the validation set of \emph{Foggy Cityscapes-refined} for $\beta=0.01$ and (b) \emph{Foggy Driving}, in terms of subjective human understanding of the foggy scenes (top) and performance of the corresponding fine-tuned DCN models (middle). For each combination of dataset and evaluation setting, we show the percentage of scenes for which each option is ranked first overall on the left, and the respective percentages for pairwise comparisons of the options on the right. Bottom: Histograms of correlation of the rankings obtained for the two evaluation settings over the datasets, measured with Kendall's $\tau$}
\label{fig:amt}
\end{figure*}

Our experiments in Section~\ref{sec:results:sl:segmentation} indicate that using any of the three examined state-of-the-art dehazing methods to preprocess foggy images before feeding them to a CNN for semantic segmentation does not provide a clear benefit over feeding the foggy images directly in the objective terms of mean IoU performance of the trained model. In this section, we complement this objective evaluation with a study of the utility of dehazing preprocessing for human understanding of foggy scenes and show that the comparative results of the objective evaluation generally agree with the comparative results of the human-based evaluation.

Both for the objective semantic-segmentation-based and the subjective human-based evaluation, we compare the four aforementioned options with regard to dehazing preprocessing \emph{individually} on each image of our datasets. \autoref{fig:dehazing:example} presents examples of the tetrads of images that we consider: the foggy image, which either belongs to the validation set of \emph{Foggy Cityscapes-refined} with $\beta=0.01$ or to \emph{Foggy Driving} and corresponds to no usage of dehazing, and its dehazed versions using DCP, MSCNN and non-local dehazing. For comparative objective evaluation of the four alternatives on each image, we use the mean IoU scores of the respective fine-tuned DCN models that are considered in the experiment of \autoref{table:finetuning:dehazing}, measured on that image. The classes that do not occur in an image are not considered for computing mean IoU on this image. The four alternatives are ranked for each image according to their mean IoU scores on it. Comparative evaluation based on human subjects considers the same tetrads of images but employs a more composite protocol, which is detailed below.

\noindent
\textbf{User Study via Amazon Mechanical Turk}. Humans are subjective and are not good at giving scores to individual images in a linear scale~\cite{paired:comparisons}. We thus follow the literature~\cite{retargetting:comparison} and choose the paired comparisons technique to let human subjects compare the four options regarding dehazing preprocessing. The participants are shown two images at a time that both pertain to the same scene, side by side, and are simply asked to choose the one which is more suitable for safe driving (\ie{}easier to interpret). Thus, six comparisons need to be performed per scene, corresponding to all possible pairs.

We use Amazon Mechanical Turk (AMT) to perform these comparisons. In order to guarantee high quality, we only employ AMT Masters in our study and verify the answers via a Known Answer Review Policy. Masters are an elite group of subjects, who have consistently demonstrated superior performance on AMT. Each individual task completed by the participants, referred to as Human Intelligence Task (HIT), comprises five image pairs to be compared, out of which three pairs are the true query pairs and the rest two pairs have a known correct answer and are only used for validation. In particular, each known-answer pair consists of two versions of a scene from \emph{Foggy Cityscapes-refined} with different levels of fog, choosing from three versions of the dataset corresponding to clear weather, $\beta=0.005$ and $\beta=0.01$. The version with less fog is considered the correct answer. In order to avoid answers based on memorized patterns, the five image pairs in each HIT are randomly shuffled and the left-right order of the images in each pair is randomly swapped. In addition, each HIT is completed by three different subjects to increase reliability. The overall quality of the user survey is shown in \autoref{fig:quality:usersurvey}, which demonstrates that the subjects have done a decent job: for $83\%$ of the HITs, both known-answer questions are answered correctly. We only use results from these HITs in our following analysis.

\noindent
\textbf{Consistency of Subjects' Answers}. 
We first study the consistency of choices among subjects; all subjects are in high agreement if the advantage of one option over the other is obvious and consistent. To measure this, we employ the coefficient of agreement~\cite{paired:comparisons}:
\begin{equation} \label{eq:agreement}
\mu = \frac{2\sigma}{\binom{m}{2}\binom{t}{2}} -1, \text{ with } \sigma=\sum_{i=1}^{t}\sum_{j=1}^{t}\binom{a_{ij}}{2},
\end{equation}
where $a_{ij}$ is the number of times that option $i$ is chosen over option $j$, $m=3$ is the number of subjects, and $t=4$ is the number of dehazing options. The maximum of $\mu$ is $1$ for complete agreement and its minimum is $-1/3$ for complete disagreement. The values of $\mu$ for all pairs of options are shown in Table~\ref{tab:agreement}. The small positive numbers in the table suggest that subjects tend to agree when comparing options pairwise but no single option has dominant advantage over another one.

\begin{table}[!tb]
  \centering
  \begin{tabular}{lr}
    \toprule
    Foggy vs. DCP & 0.155\\
    Foggy vs. MSCNN & 0.115\\
    Foggy vs. Non-local & 0.010\\
    DCP vs. MSCNN & 0.182\\
    DCP vs. Non-local & 0.036\\
    MSCNN vs. Non-local & 0.182\\
    \midrule
    Mean & 0.113\\
    \bottomrule
  \end{tabular}
  \caption{Agreement coefficients for all pairwise comparisons of the four dehazing options}
  \label{tab:agreement}
\end{table}

\noindent
\textbf{Ranking and Correlation with Objective Evaluation}. We finally compute the overall ranking of all four options for each image based on the number of times each option is chosen in all relevant pairwise comparisons. The correlation of these rankings with those induced by mean IoU performance is measured with \emph{Kendall's $\tau$ coefficient}~\cite{kendall1938new} with $-1 \leq \tau \leq 1$, where a value of $1$ implies perfect agreement, $-1$ implies perfect disagreement, and $0$ implies zero correlation. \autoref{fig:amt} provides a complete overview of the comparative results both for our user study and the semantic-segmentation-based evaluation on \emph{Foggy Cityscapes-refined} and \emph{Foggy Driving}, including rank correlation results for the two types of evaluation.

The results in the top row of \autoref{fig:amt} indicate that none of the three examined methods for dehazing preprocessing improves reliably the human understanding of synthetic foggy scenes from \emph{Foggy Cityscapes} or real foggy scenes from \emph{Foggy Driving}. In particular, the no-dehazing option beats all other three options in pairwise comparisons on \emph{Foggy Cityscapes-refined} and loses only to DCP marginally on \emph{Foggy Driving}, while it is also ranked first on more images than any other option for both datasets.

In addition, the rankings obtained with the two types of evaluation are generally in congruence for the real-world case of \emph{Foggy Driving}. The no-dehazing and DCP options are ranked higher than MSCNN and non-local dehazing both in the user study and in the objective evaluation. The high performance of DCP compared to MSCNN is due to the usage of $\beta=0.01$ for \emph{Foggy Cityscapes-refined} (cf.\ the discussion in Section~\ref{sec:results:sl:segmentation}). What is more, the two rankings exhibit a positive correlation on average for \emph{Foggy Driving} based on the respective distribution of $\tau$ in the bottom right chart of \autoref{fig:amt}, which supports our conclusion in Section~\ref{sec:results:sl:segmentation} about the marginal benefit of dehazing preprocessing for foggy scene understanding.

\subsection{Object Detection}
\label{sec:results:sl:detection}

For our experiment on object detection in foggy scenes, we select the modern Fast R-CNN~\cite{fast:rcnn} as the architecture of the evaluated models. We prefer Fast R-CNN over more recent approaches such as Faster R-CNN~\cite{faster:rcnn} because the former involves a simpler training pipeline, making fine-tuning to foggy conditions straightforward. Consequently, we do not learn the front-end of the object detection pipeline which involves generation of object proposals; rather, we use multiscale combinatorial grouping~\cite{mcg} for this task.

In order to ensure a fair comparison, we first obtain a baseline Fast R-CNN model for the original Cityscapes dataset, similarly to the preceding semantic segmentation experiments. Since no such model is publicly available, we begin with the model released by the author of~\cite{fast:rcnn} which has been trained on PASCAL VOC 2007~\cite{pascal:2011} and fine-tune it on the union of the training and validation sets of Cityscapes which comprises 3475 images. Fine-tuning through all layers is run with the same configurations as in~\cite{fast:rcnn}, except that we use the ``poly'' learning rate policy with a base learning rate of $2\times{}10^{-4}$ and a power parameter of $0.9$, with 7k iterations (4 epochs).

This baseline model that has been trained on the real Cityscapes with clear weather serves as initialization for fine-tuning on our synthetic images from \emph{Foggy Cityscapes-refined}. To this end, we use all 550 training and validation images of \emph{Foggy Cityscapes-refined} and fine-tune with the same settings as before, only that the base learning rate is set to $10^{-4}$ and we run 1650 iterations (6 epochs).

We experiment with two values of the attenuation coefficient $\beta$ for \emph{Foggy Cityscapes-refined} and present comparative performance on the 33 finely annotated images of \emph{Foggy Driving} in \autoref{table:detection}. No dehazing is involved in this experiment. We concentrate on the classes \emph{car} and \emph{person} for evaluation, since they constitute the intersection of the set of frequent classes in \emph{Foggy Driving} and the set of annotated classes with distinct instances. Individual average precision (AP) scores for \emph{car} and \emph{person} are reported, as well as mean scores over these two classes (``mean frequent'') and over the complete set of 8 classes occurring in instances (``mean all''). For completeness, we note that the original VOC 2007 model of~\cite{fast:rcnn} exhibits an AP of 2.1\% for \emph{car} and 1.9\% for \emph{person}.

\begin{figure*}[tb]
    \centering
    \subfloat{\includegraphics[width=0.32\textwidth]{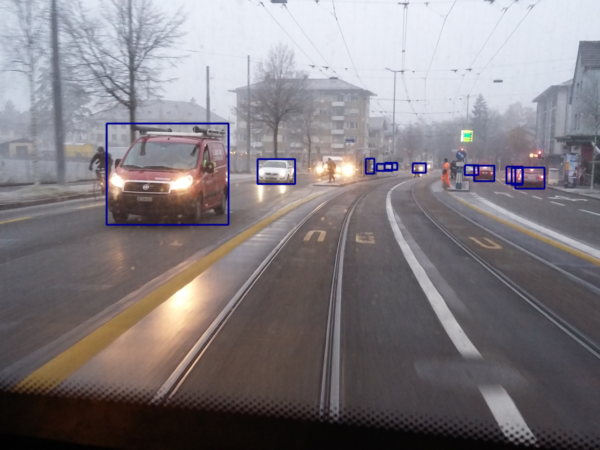}}
    \hfil
    \subfloat{\includegraphics[width=0.32\textwidth]{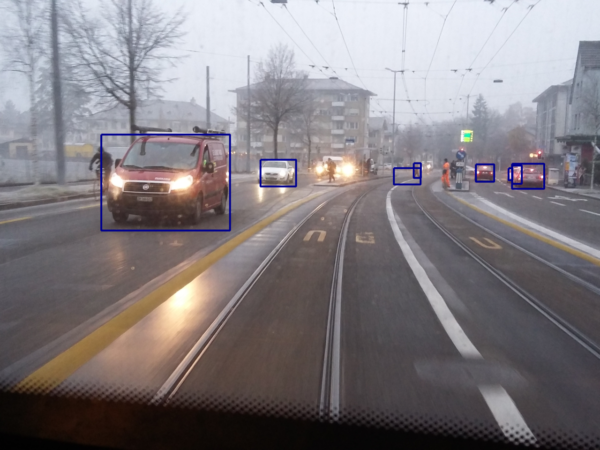}}
    \hfil
    \subfloat{\includegraphics[width=0.32\textwidth]{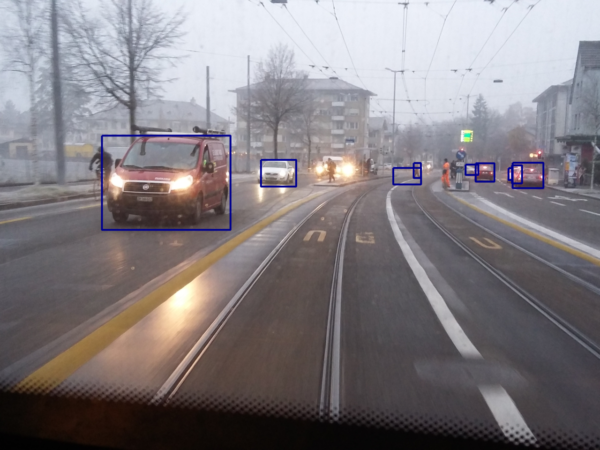}}
    \\
    \subfloat{\includegraphics[width=0.32\textwidth]{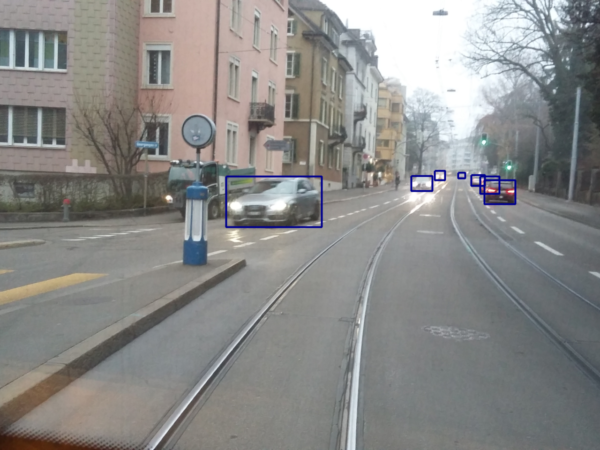}}
    \hfil
    \subfloat{\includegraphics[width=0.32\textwidth]{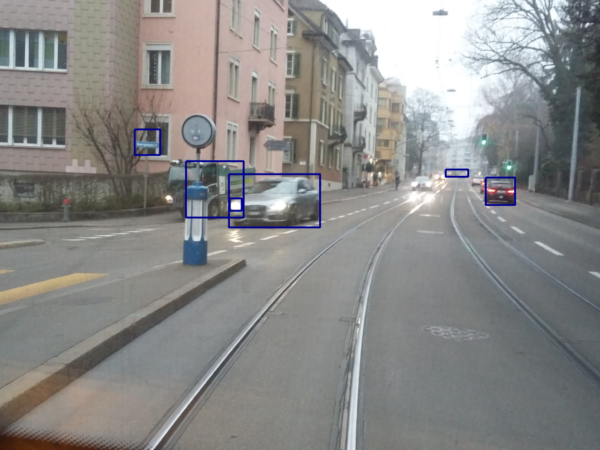}}
    \hfil
    \subfloat{\includegraphics[width=0.32\textwidth]{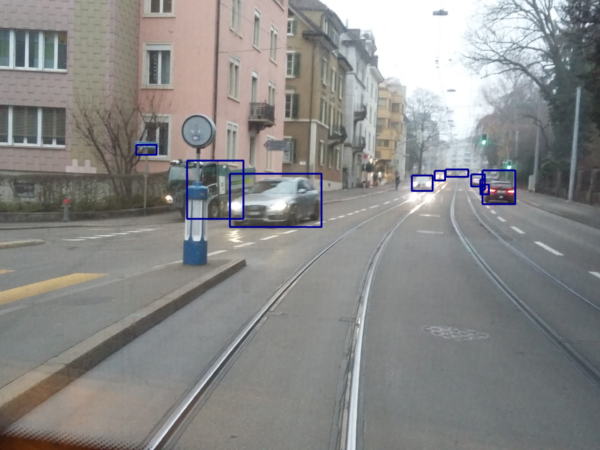}}
    \\
    \subfloat{\includegraphics[width=0.32\textwidth]{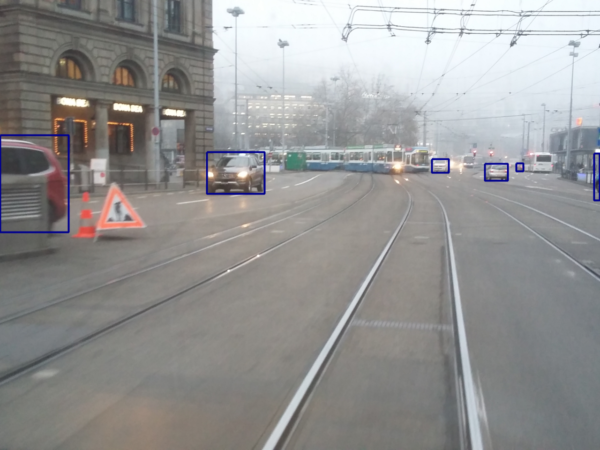}}
    \hfil
    \subfloat{\includegraphics[width=0.32\textwidth]{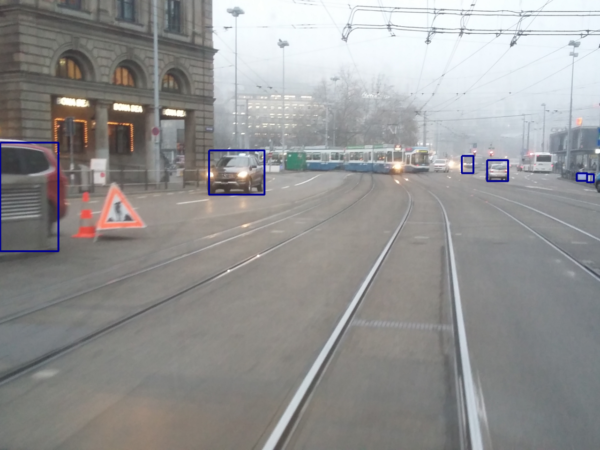}}
    \hfil
    \subfloat{\includegraphics[width=0.32\textwidth]{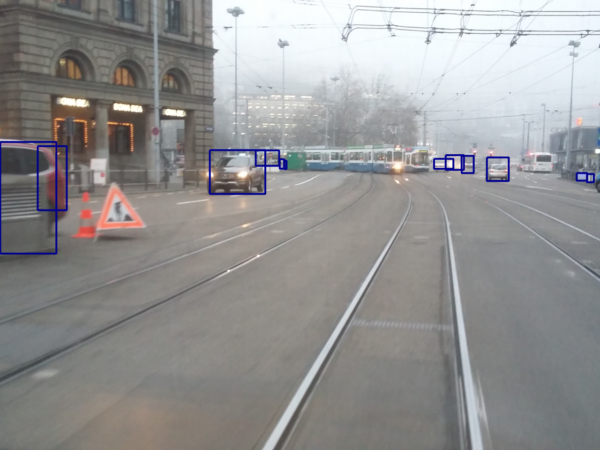}}
    \caption{Qualitative results for detection of cars on \emph{Foggy Driving}. From left to right: ground truth annotation, baseline Fast R-CNN model trained on original Cityscapes, and our model \emph{FT-0.005} fine-tuned on \emph{Foggy Cityscapes-refined} with light fog. This figure is seen better when zoomed in on a screen}
    \label{fig:detection:results}
\end{figure*}

\begin{table}[!tb]
  \centering
  \caption{Performance comparison on \emph{Foggy Driving} of baseline Fast R-CNN model trained on Cityscapes (``W/o FT'') versus fine-tuned versions of it using \emph{Foggy Cityscapes-refined}. ``FT'' stands for using fine-tuning and ``W/o FT'' for not using fine-tuning. AP (\%) is used to report results}
  \label{table:detection}
  \setlength\tabcolsep{4pt}
  \begin{tabular*}{\linewidth}{l @{\extracolsep{\fill}} cccc}
  \toprule
  & mean all & car & person & mean frequent\\
  \midrule
  W/o FT & 11.1 & 30.5 & \best{10.3} & 20.4\\
  FT $\beta=0.01$ & 11.1 & 34.6 & 10.0 & 22.3\\
  FT $\beta=0.005$ & \best{11.7} & \best{35.3} & \best{10.3} & \best{22.8}\\
  \bottomrule
  \end{tabular*}
\end{table}

Both of our fine-tuned models outperform the baseline model by a significant margin for \emph{car}. At the same time, they are on a par with the baseline model for \emph{person}. The overall winner is the model that has been fine-tuned on light fog, which we refer to as \emph{FT-0.005}: it outperforms the baseline model by 2.4\% on average on the two frequent classes and it is also slightly better when taking all 8 classes into account.

We provide a visual comparison of \emph{FT-0.005} and the baseline model for car detection on example images from \emph{Foggy Driving} in \autoref{fig:detection:results}. Note the ability of our model to detect distant cars, such as the two cars in the image of the second row which are moving on the left side of the road and are visible from their front part. These two cars are both missed by the baseline model.

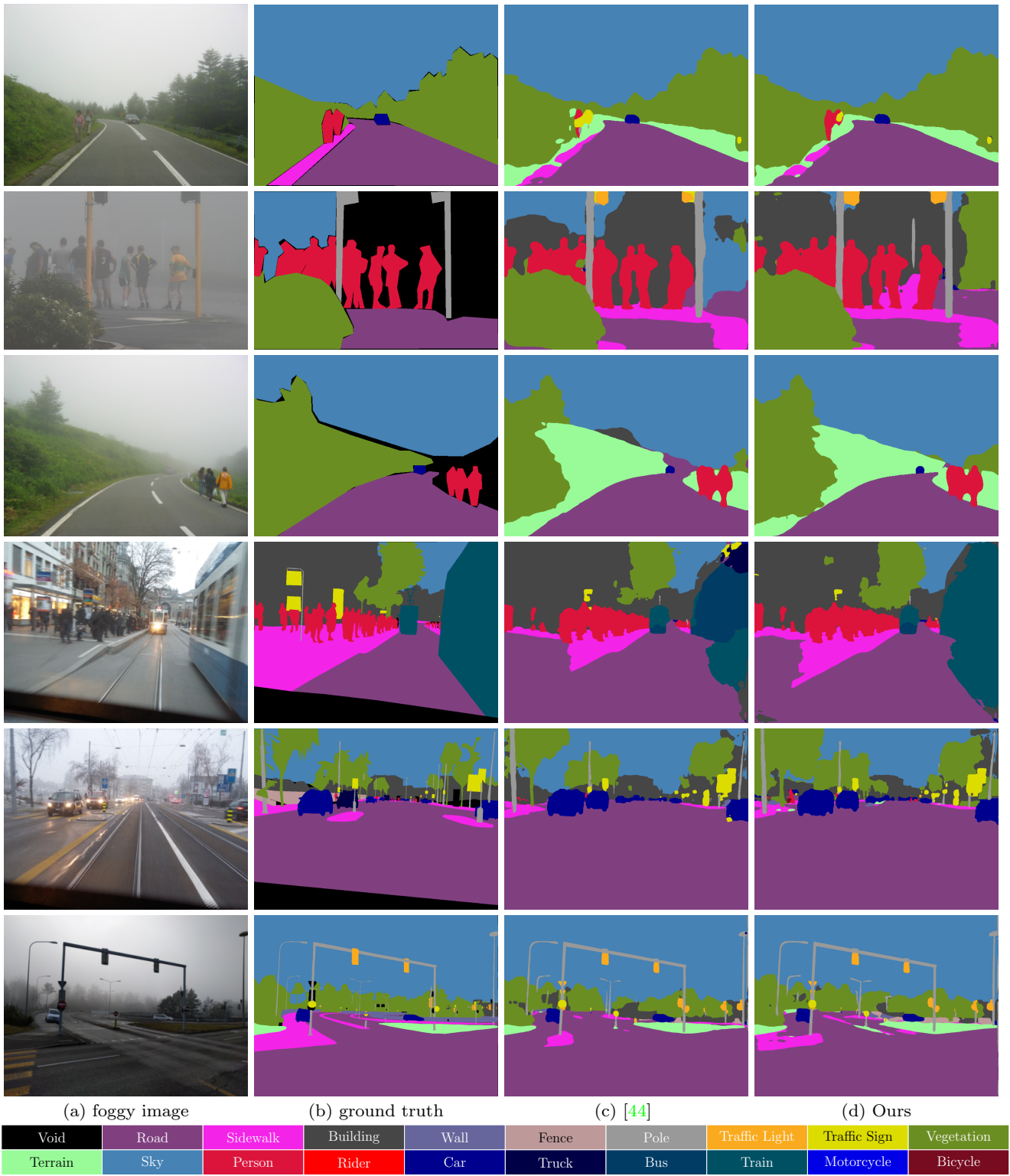
\begin{figure*}[!tb]
\centering
\begin{tabular}{ccccc}
\hspace{-2mm}
\includegraphics[width=0.24\textwidth]{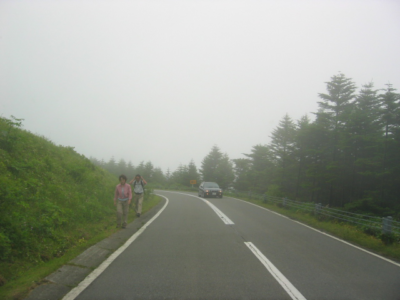} 
& \hspace{-4mm}
\includegraphics[width=0.24\textwidth]{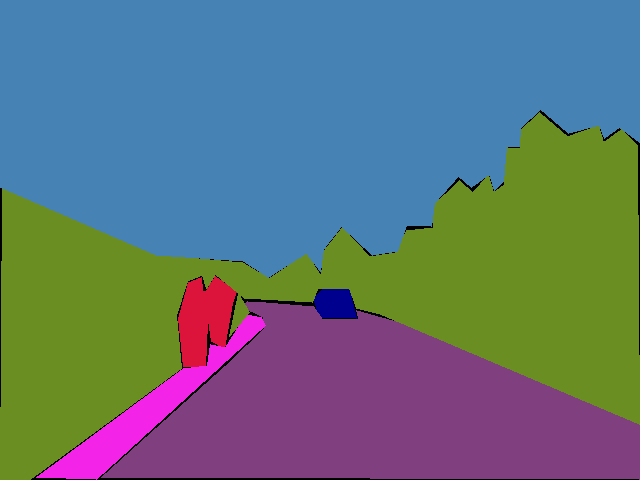}  
& \hspace{-4mm}   
\includegraphics[width=0.24\textwidth]{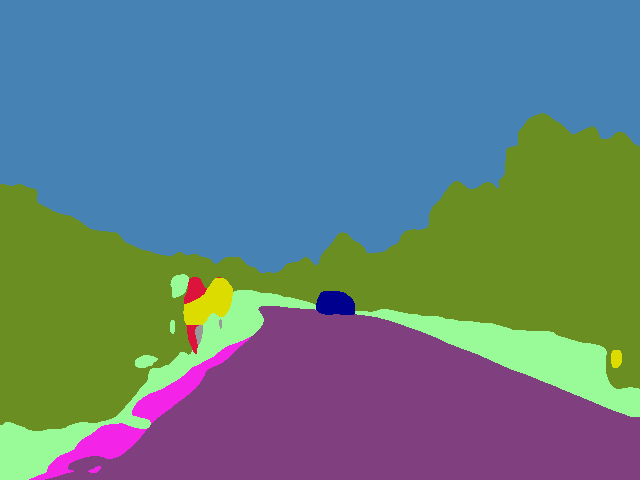} 
& \hspace{-4mm} 
\includegraphics[width=0.24\textwidth]{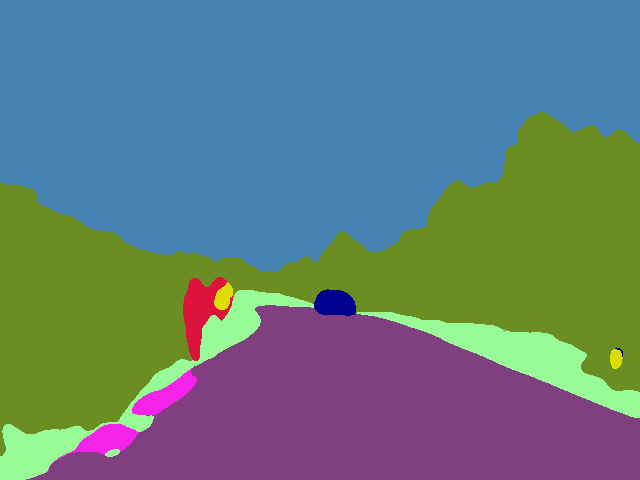} \\ 
\hspace{-2mm}
\includegraphics[width=0.24\textwidth]{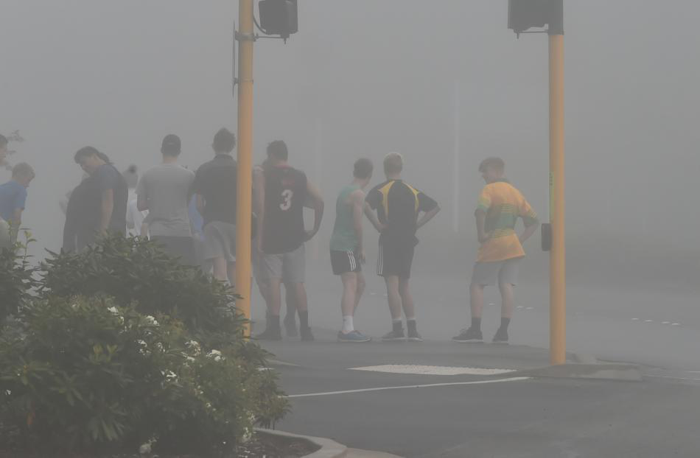} 
& \hspace{-4mm}
\includegraphics[width=0.24\textwidth]{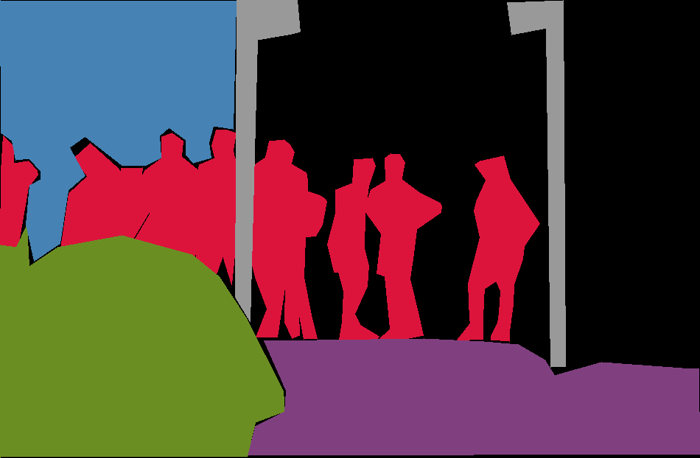}  
& \hspace{-4mm}   
\includegraphics[width=0.24\textwidth]{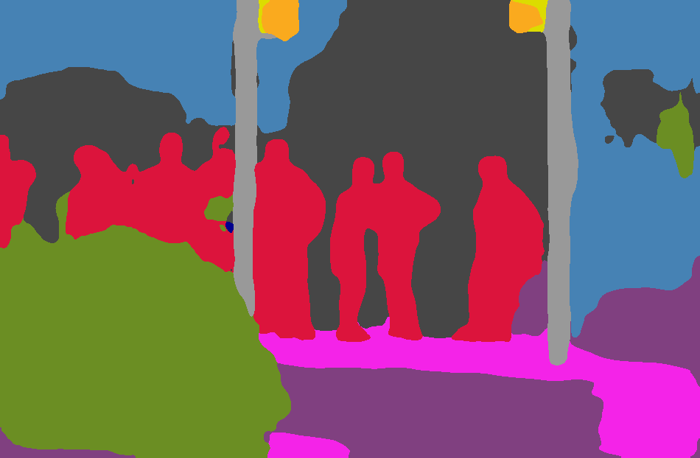} 
& \hspace{-4mm} 
\includegraphics[width=0.24\textwidth]{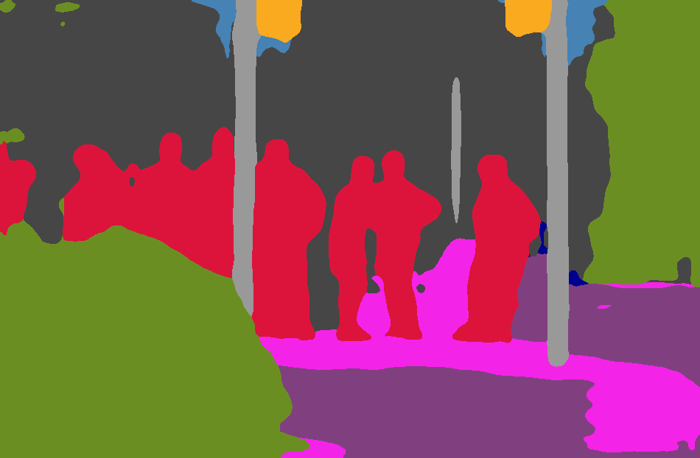} \\ 
\hspace{-2mm}
\includegraphics[width=0.24\textwidth]{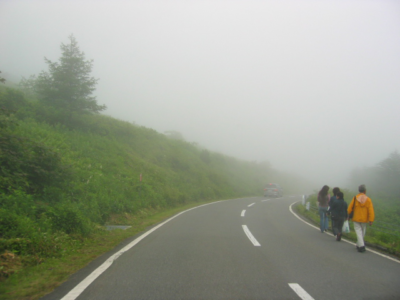} 
& \hspace{-4mm}
\includegraphics[width=0.24\textwidth]{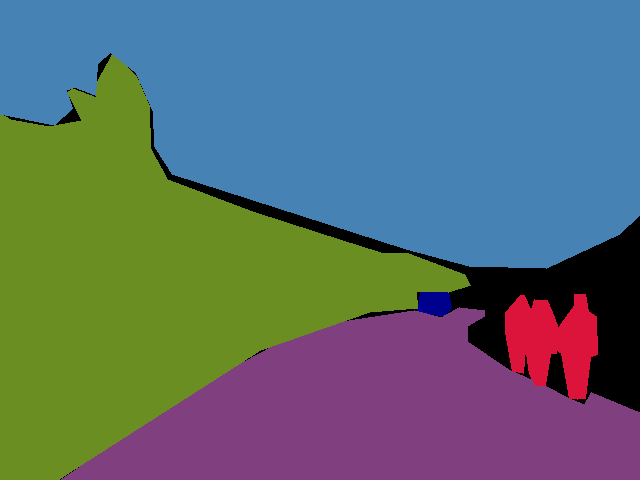}  
& \hspace{-4mm}   
\includegraphics[width=0.24\textwidth]{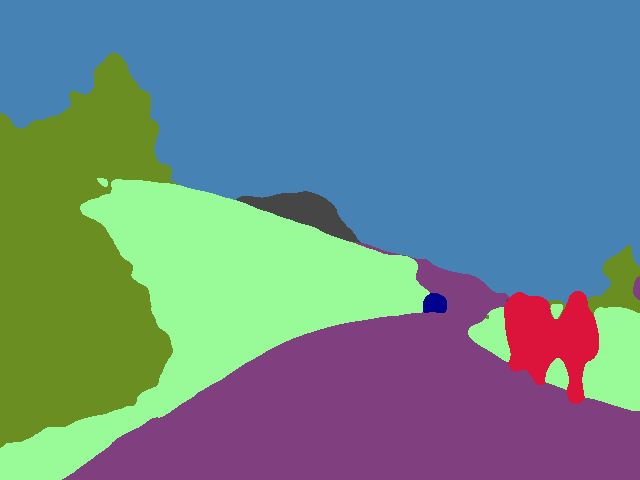} 
& \hspace{-4mm} 
\includegraphics[width=0.24\textwidth]{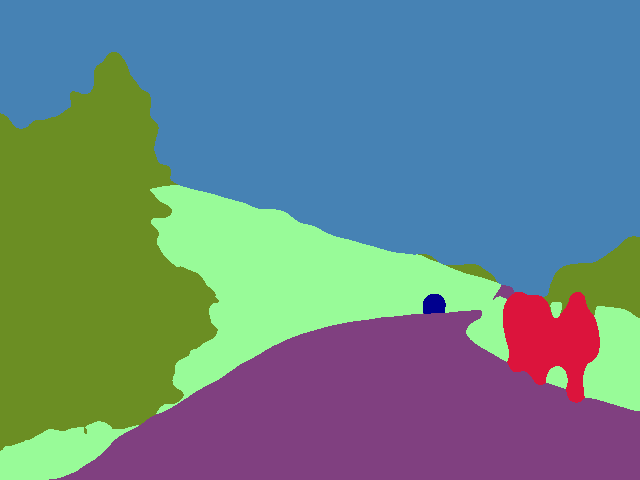} \\ 
\hspace{-2mm}
\includegraphics[width=0.24\textwidth]{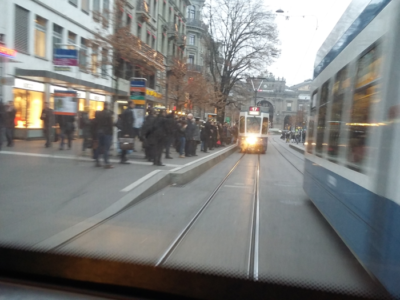} 
& \hspace{-4mm}
\includegraphics[width=0.24\textwidth]{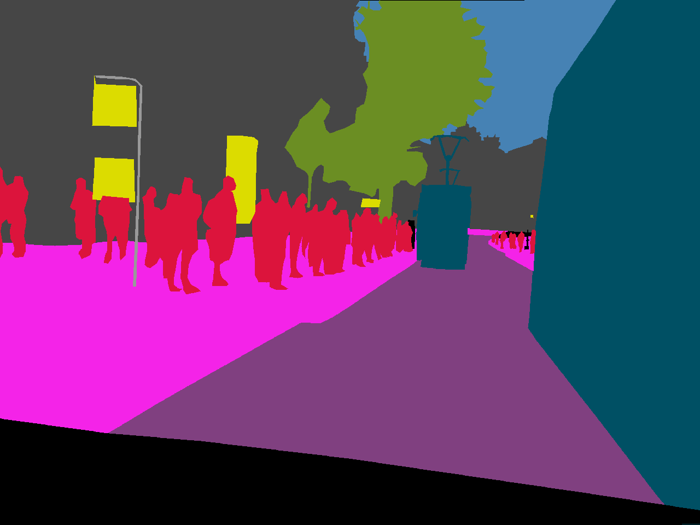}  
& \hspace{-4mm}   
\includegraphics[width=0.24\textwidth]{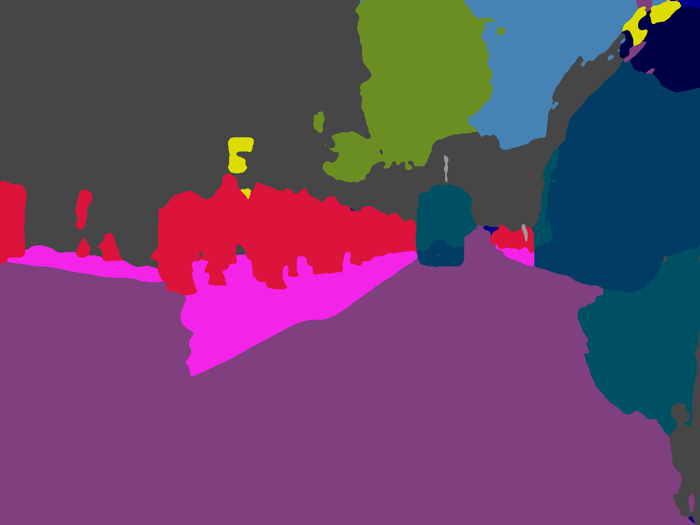} 
& \hspace{-4mm} 
\includegraphics[width=0.24\textwidth]{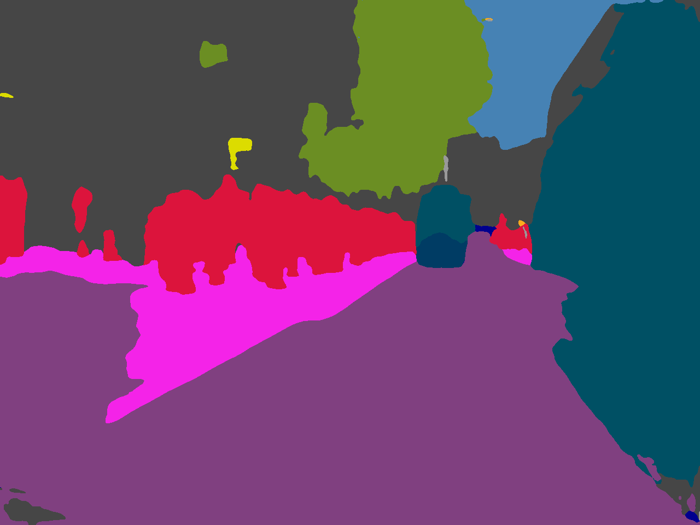} \\ 
\hspace{-2mm}
\includegraphics[width=0.24\textwidth]{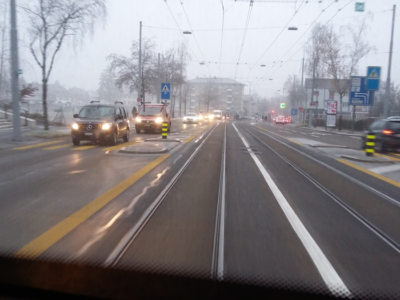} 
& \hspace{-4mm}
\includegraphics[width=0.24\textwidth]{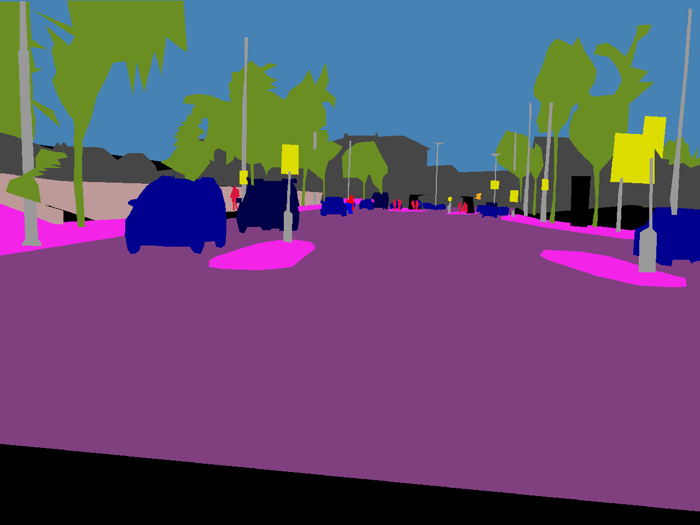}  
& \hspace{-4mm}   
\includegraphics[width=0.24\textwidth]{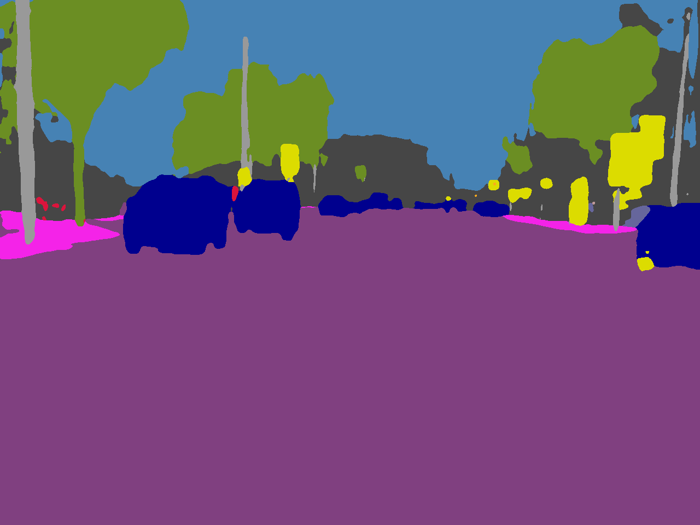} 
& \hspace{-4mm} 
\includegraphics[width=0.24\textwidth]{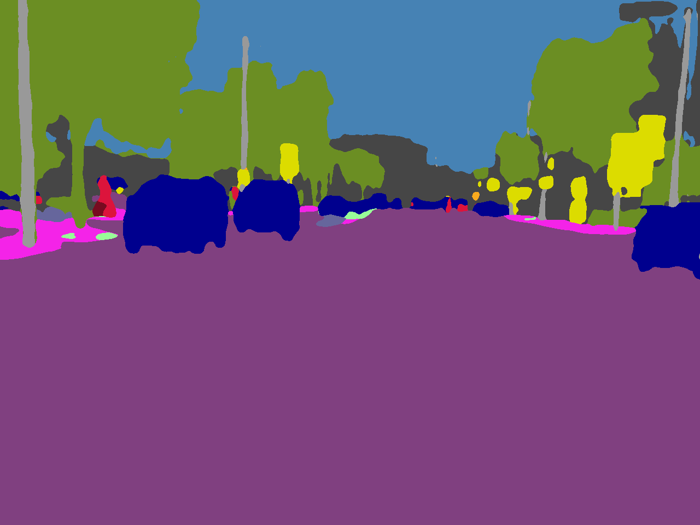} \\ 
\hspace{-2mm}
\includegraphics[width=0.24\textwidth]{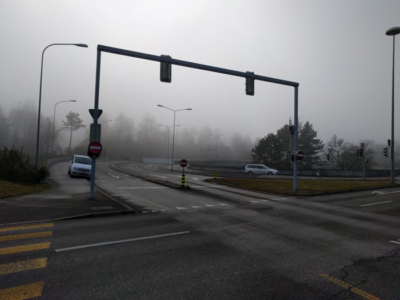} 
& \hspace{-4mm}
\includegraphics[width=0.24\textwidth]{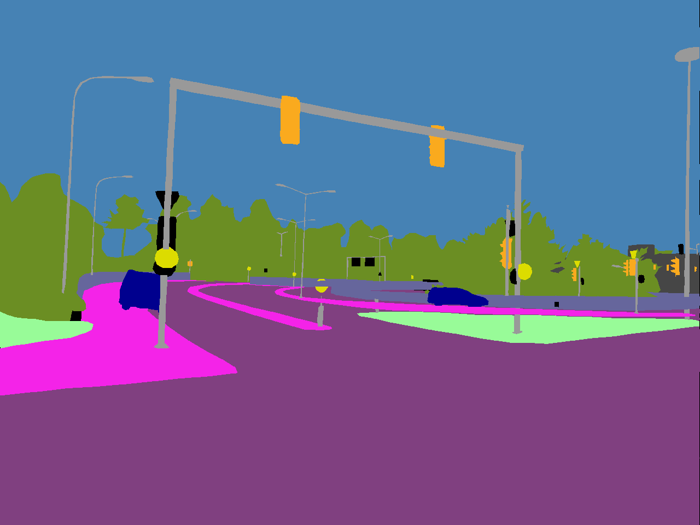}  
& \hspace{-4mm}   
\includegraphics[width=0.24\textwidth]{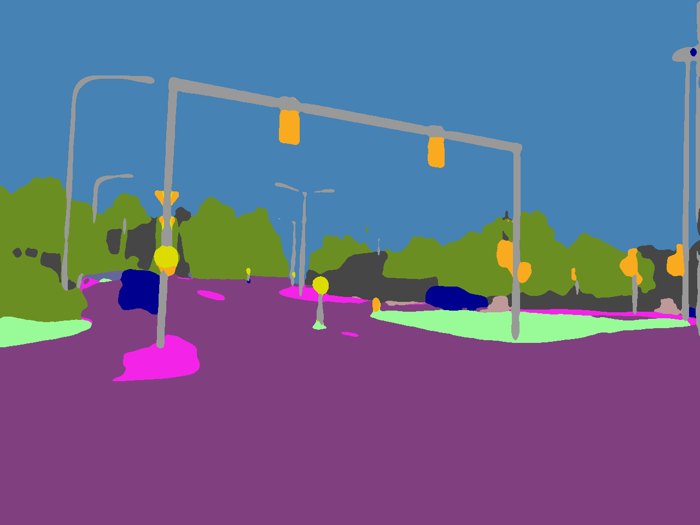} 
& \hspace{-4mm} 
\includegraphics[width=0.24\textwidth]{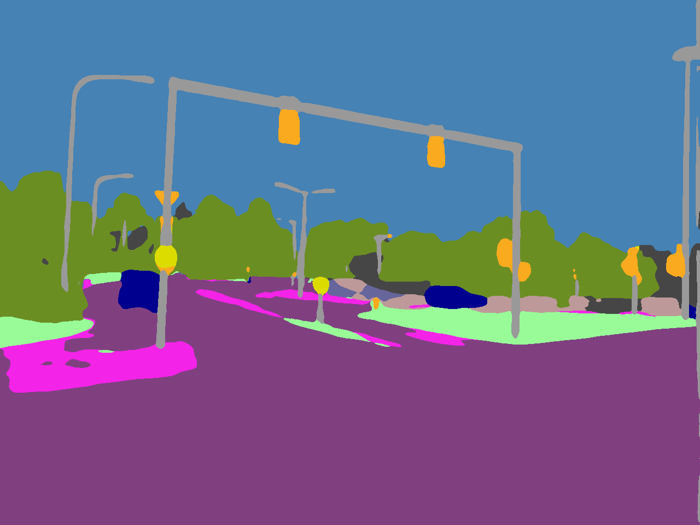} \\ 
\hspace{-2mm} \text{(a) foggy image} &  \hspace{-4mm} (b) ground truth &  \hspace{-4mm} \text{(c) \cite{refinenet} } &  \hspace{-4mm} \text{(d) Ours}  \\
\end{tabular}
\resizebox{\linewidth}{!}{
\begin{tikzpicture}[tight background, scale=0.75, every node/.style={font=\large}]
	\draw[white, fill=void, draw=white] (0,0) rectangle (1 * 4, 1) node[pos=0.5] {Void};
	\draw[white, fill=road, draw=white] (1 * 4,0) rectangle (2 * 4, 1) node[pos=0.5] {Road};
	\draw[white, fill=sidewalk, draw=white] (2 * 4,0) rectangle (3 * 4, 1) node[pos=0.5] {Sidewalk};
	\draw[white, fill=building, draw=white] (3 * 4,0) rectangle (4 * 4, 1) node[pos=0.5] {Building};
	\draw[white, fill=wall, draw=white] (4 * 4,-0) rectangle (5 * 4, 1) node[pos=0.5] {Wall};
	\draw[black, fill=fence, draw=white] (5 * 4,-0) rectangle (6 * 4, 1) node[pos=0.5] {Fence};
	\draw[white, fill=pole, draw=white] (6 * 4,-0) rectangle (7 * 4, 1) node[pos=0.5] {Pole};
	\draw[white, fill=traffic light, draw=white] (7 * 4,-0) rectangle (8 * 4, 1) node[pos=0.5] {Traffic Light};
	\draw[black, fill=traffic sign, draw=white] (8 * 4,-0) rectangle (9 * 4, 1) node[pos=0.5] {Traffic Sign};
	\draw[white, fill=vegetation, draw=white] (9 * 4,-0) rectangle (10 * 4, 1) node[pos=0.5] {Vegetation};

	\draw[black, fill=terrain, draw=white] (0 * 4,-1) rectangle (1 * 4, 0) node[pos=0.5] {Terrain};
	\draw[white, fill=sky, draw=white] (1 * 4,-1) rectangle (4 * 2, 0) node[pos=0.5] {Sky};
	\draw[white, fill=person, draw=white] (2 * 4,-1) rectangle (3 * 4, 0) node[pos=0.5] {Person};
	\draw[white, fill=rider, draw=white] (3 * 4,-1) rectangle (4 * 4, 0) node[pos=0.5] {Rider};
	\draw[white, fill=car, draw=white] (4 * 4,-1) rectangle (5 * 4, 0) node[pos=0.5] {Car};
	\draw[white, fill=truck, draw=white] (5 * 4,-1) rectangle (6 * 4, 0) node[pos=0.5] {Truck};
	\draw[white, fill=bus, draw=white] (6 * 4,-1) rectangle (7 * 4, 0) node[pos=0.5] {Bus};
	\draw[white, fill=train, draw=white] (7 * 4,-1) rectangle (8 * 4, 0) node[pos=0.5] {Train};
	\draw[white, fill=motorcycle, draw=white] (8 * 4,-1) rectangle (9 * 4, 0) node[pos=0.5] {Motorcycle};
	\draw[white, fill=bicycle, draw=white] (9 * 4,-1) rectangle (10 * 4, 0) node[pos=0.5] {Bicycle};
\end{tikzpicture}}
\caption{Qualitative results for semantic segmentation on \emph{Foggy Driving}, both for coarsely annotated images (top three rows) and finely annotated images (bottom three rows). ``Ours'' stands for \cite{refinenet} fine-tuned with our SSL on \emph{Foggy Cityscapes}}
\label{fig:sem:seg}
\end{figure*}

\section{Semi-supervised Learning with Synthetic Fog}
\label{sec:learning:ssl}
While standard supervised learning can improve the performance of SFSU using our synthetic fog, the paradigm still needs manual annotations for corresponding clear-weather images.  
In this section, we extend the learning to a new paradigm which is also able to acquire knowledge from unlabeled pairs of foggy images and clear-weather images. In particular, we train a semantic segmentation model on clear-weather images using the standard supervised learning paradigm, and apply the model to an even larger set of clear but ``unlabeled'' images (\eg{}our 20000 unlabeled images of \emph{Foggy Cityscapes-coarse}) to generate the class responses. Since we have created a foggy version for the unlabeled dataset, these class responses can then be used to supervise the training of models for SFSU. 

This learning approach is inspired by the stream of work on model distillation~\cite{hinton2015distilling,supervision:transfer} or imitation~\cite{model:compression,dai:metric:imitation}. \cite{model:compression,hinton2015distilling,dai:metric:imitation} transfer supervision from sophisticated models to simpler models for efficiency, and \cite{supervision:transfer} transfers supervision from the domain of images to other domains such as depth maps. In our case, supervision is transferred from clear weather to foggy weather.
The underpinnings of our proposed approach are the following: 1) in clear weather, objects are easier to recognize than in foggy weather, thus models trained on images with clear weather in principle generalize better to new images of the same condition than those trained on foggy images; and 2) since the synthetic foggy images and their clear-weather counterparts depict exactly the same scene, recognition results should also be the same for both images ideally.

We formulate our semi-supervised learning (SSL) for semantic segmentation as follows. Let us denote a clear-weather image by $\mathbf{x}$, the corresponding foggy one by $\mathbf{x}^\prime$, and the corresponding human annotation by $\mathbf{y}$. Then, the training data consist of both labeled data $\mathcal{D}_l =\{(\mathbf{x}_i, \mathbf{x}^\prime_i, \mathbf{y}_i)\}_{i=1}^{l}$ and unlabeled data $\mathcal{D}_u =\{(\mathbf{x}_j, \mathbf{x}^\prime_j)\}_{j=l+1}^{l+u}$, where $\mathbf{y}_i^{m,n} \in\{1, ..., K\}$  is the label of pixel $(m,n)$, and $K$ is the total number of classes. $l$ is the number of labeled training images, and $u$ is the number of unlabeled training images. The aim is to learn a mapping function  $\phi^\prime: \mathcal{X}^\prime \mapsto \mathcal{Y}$ from $\mathcal{D}_l$ and $\mathcal{D}_u$. In our case, $\mathcal{D}_l$ consists of the 498 high-quality foggy images in the training set of \emph{Foggy Cityscapes-refined} which have human annotations with fine details, and $\mathcal{D}_u$ consists of the additional $20000$ foggy images in \emph{Foggy Cityscapes-coarse} which do not have fine human annotations. 

Since $\mathcal{D}_u$ does not have class labels, we use the idea of supervision transfer to generate the supervisory labels for all the images therein.  To this end, we first learn a mapping function $\phi: \mathcal{X} \mapsto \mathcal{Y}$ with $\mathcal{D}_l$ and then obtain the labels $\hat{\mathbf{y}}_j=\phi(\mathbf{x}_j)$ for $\mathbf{x}_j$ and $\mathbf{x}^\prime_j$, $\forall j \in \{l+1, ..., l+u\}$.  $\mathcal{D}_u$ is then upgraded to $\hat{\mathcal{D}}_u=\{(\mathbf{x}_j, \mathbf{x}^\prime_j, \hat{\mathbf{y}}_j)\}_{j=l+1}^{l+u}$. The proposed  scheme for training semantic segmentation models for foggy images $\mathbf{x}^\prime$ is to learn a mapping function  $\phi^\prime$ so that  human annotations $\mathbf{y}$ and the transferred labels $\hat{\mathbf{y}}$ are both taken into account: 
\begin{equation}
\min_{\phi^\prime} \sum_{i=1}^l L(\phi^\prime(\mathbf{x}^\prime_i), \mathbf{y}_i) + \lambda \sum_{j=l+1}^{l+u} L(\phi^\prime(\mathbf{x}^\prime_j), \hat{\mathbf{y}}_j),
\label{eq:ssl}
\end{equation}
where $L(.,.)$ is the Categorical Cross Entropy Loss function for classification, and $\lambda=\frac{l}{u}\times w$ is a parameter for balancing the contribution of the two terms, serving as the relative weight of each unlabeled image compared to each labeled one. We empirically set $w=5$ in our experiment, but an optimal value can be obtained via cross-validation if needed. In our implementation, we approximate the optimization of \eqref{eq:ssl} by mixing images from $\mathcal{D}_l$ and $\hat{\mathcal{D}}_u$ in a proportion of $1:w$ and feeding the stream of hybrid data to a CNN for standard supervised training.

We select RefineNet~\cite{refinenet} as the CNN model for semantic segmentation, which is a more recent and better performing method than DCN~\cite{dilated:convolution} that is used in Section~\ref{sec:learning:sl}. The reason for using DCN in Section~\ref{sec:learning:sl} is that RefineNet had not been published yet at the time that we were conducting the experiments of Section~\ref{sec:learning:sl}. We would like to note that the state-of-the-art PSPNet~\cite{pspnet}, which has been trained on the Cityscapes dataset similarly to the original version of RefineNet that we use as our baseline, achieved a mean IoU of only 24.0\% on \emph{Foggy Driving} in our initial experiments.

We use mean IoU for evaluation, similarly to Section~\ref{sec:learning:sl}, and $\beta=0.01$ for \emph{Foggy Cityscapes}. We compare the performance of three trained models: 1) original RefineNet~\cite{refinenet} trained on Cityscapes, 2) RefineNet fine-tuned on $\mathcal{D}_l$, and 3) RefineNet fine-tuned on $\mathcal{D}_l$ and $\hat{\mathcal{D}}_u$. The mean IoU scores of the three models on \emph{Foggy Driving} are $44.3\%$, $46.3\%$, and $\mathbf{49.7\%}$ respectively. The $2\%$ improvement of 2) over 1) confirms the conclusion we draw in Section~\ref{sec:learning:sl} that fine-tuning with our synthetic fog can indeed improve the performance of semantic foggy scene understanding. The $3.4\%$ improvement of 3) over 2) validates the efficacy of the SSL paradigm.  \autoref{fig:sem:seg} shows visual results of 1) and 3), along with the foggy images and human annotations. The re-trained model with our SSL paradigm can better segment certain parts of the images which are misclassified by the original RefineNet, \eg{}the pedestrian in the first example, the tram in the fourth one, and the sidewalk in the last one.

Both the quantitative and qualitative results suggest that our approach is able to alleviate the need for collecting large-scale training data for semantic understanding of foggy scenes, by training with the  annotations that are already available for clear-weather images and the generated foggy images directly and by transferring supervision from clear-weather images to foggy images of the same scenes.

\section{Conclusion}
\label{sec:conclusion}

In this paper, we have demonstrated the benefit of synthetic data that are based on real images for semantic understanding of foggy scenes. Two foggy datasets have been constructed to this end: the partially synthetic \emph{Foggy Cityscapes} dataset which derives from Cityscapes, and the real-world \emph{Foggy Driving} dataset, both with dense pixel-level semantic annotations for 19 classes and bounding box annotations for objects belonging to 8 classes. We have shown that \emph{Foggy Cityscapes} can be used to boost performance of state-of-the-art CNN models for semantic segmentation and object detection on the challenging real foggy scenes of \emph{Foggy Driving}, both in a usual supervised setting and in a novel, semi-supervised setting. Last but not least, we have exposed through detailed experiments the fact that image dehazing faces difficulties in working out of the box on real outdoor foggy data and thus is marginally helpful for SFSU. In the future, we would like to combine dehazing and semantic understanding of foggy scenes into a unified, end-to-end learned pipeline, which can also leverage the type of synthetic foggy data we have introduced. The datasets, models and code are available at \url{http://www.vision.ee.ethz.ch/~csakarid/SFSU_synthetic}.

\begin{acknowledgements}
The authors would like to thank Kevis Maninis for useful discussions. This work is funded by Toyota Motor Europe via the research project TRACE-Zürich.
\end{acknowledgements}

\bibliographystyle{spmpsci}      
\bibliography{references_ijcv.bib}

\end{document}